\documentclass[letterpaper]{article} 
\usepackage{aaai25}  
\usepackage{times}  
\usepackage{helvet}  
\usepackage{courier}  
\usepackage[hyphens]{url}  
\usepackage{graphicx} 
\usepackage{natbib}  
\usepackage{caption} 
\usepackage{algorithm}
\usepackage{algorithmic}

\usepackage{amsfonts}
\usepackage{amsmath}
\usepackage{bigstrut,multirow,rotating,booktabs}
\usepackage{xcolor}
\usepackage{cuted}

\usepackage{newfloat}
\usepackage{listings}
\DeclareCaptionStyle{ruled}{labelfont=normalfont,labelsep=colon,strut=off} 
\floatstyle{ruled}
\newfloat{listing}{tb}{lst}{}
\floatname{listing}{Listing}
\title{Kernel-Aware Graph Prompt Learning for Few-Shot Anomaly Detection}
\author {
    Fenfang Tao\textsuperscript{\rm 1},
    Guo-Sen Xie\textsuperscript{\rm 1}\thanks{Corresponding Author}, 
    Fang Zhao\textsuperscript{\rm 2},
    Xiangbo Shu\textsuperscript{\rm 1}
} 
\affiliations {
    \textsuperscript{\rm 1}School of Computer Science and Engineering, Nanjing University of Science and Technology, Nanjing, China\\
    \textsuperscript{\rm 2}School of Intelligence Science and Technology, Nanjing University, Suzhou, China\\
    \{fenfangtao294, gsxiehm, zhaofang0627, shuxb104\}@gmail
}

\usepackage{bibentry}

\begin{document}

\maketitle
\begin{abstract}
Few-shot anomaly detection (FSAD) aims to detect unseen anomaly regions with the guidance of very few normal support images from the same class. Existing FSAD methods usually find anomalies by directly designing complex text prompts to align them with visual features under the prevailing large vision-language model paradigm. However, these methods, almost always, neglect intrinsic contextual information in visual features, e.g., the interaction relationships between different vision layers, which is an important clue for detecting anomalies comprehensively. To this end, we propose a kernel-aware graph prompt learning framework, termed as KAG-prompt, by reasoning the cross-layer relations among visual features for FSAD. Specifically, a kernel-aware hierarchical graph is built by taking the different layer features focusing on anomalous regions of different sizes as nodes, meanwhile, the relationships between arbitrary pairs of nodes stand for the edges of the graph. By message passing over this graph, KAG-prompt can capture cross-layer contextual information, thus leading to more accurate anomaly prediction. Moreover, to integrate the information of multiple important anomaly signals in the prediction map, we propose a novel image-level scoring method based on multi-level information fusion. Extensive experiments on MVTecAD and VisA datasets show that KAG-prompt achieves state-of-the-art FSAD results for image-level/pixel-level anomaly detection. Code is available at https://github.com/CVL-hub/KAG-prompt.git.
\end{abstract}

%

\begin{figure}[ht!]
		\centering
		\includegraphics[width=1\linewidth]{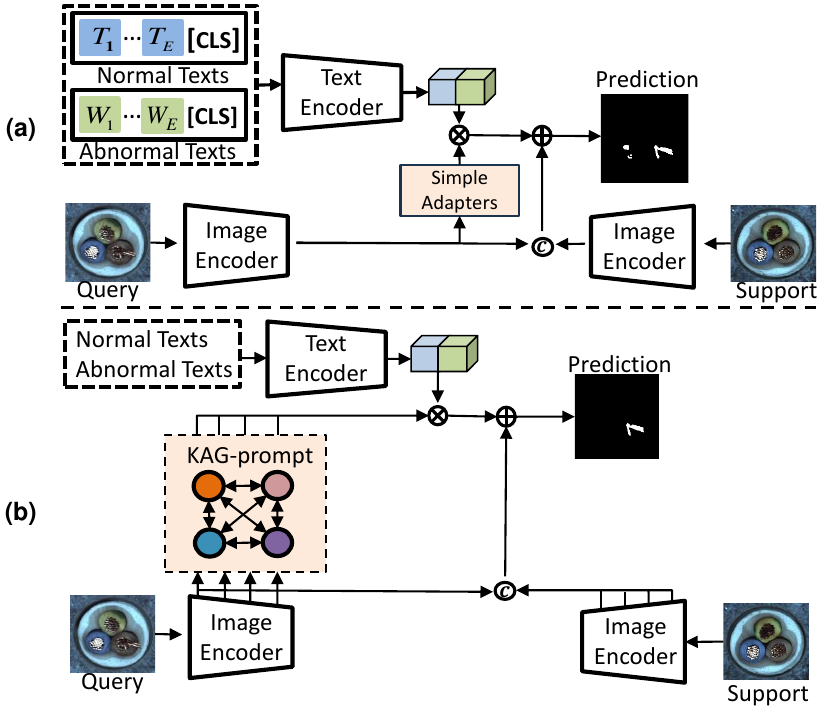}
		\caption{\textbf{Comparisons of KAG-prompt and existing FSAD models.} (a) Existing FSAD methods usually design complex text prompts, i.e., $T_{i}, W_{i}$ are manually designed and/or learnable text prompts. For query image branches, they only learn simple adapters to extract visual features for downstream tasks. However, this paradigm segments normal backgrounds into anomaly ones. (b) Our KAG-prompt can well predict the anomalies in the query image by constructing a kernel-aware hierarchical graph to capture cross-layer multi-level relationships.}
		\label{fig1}
\end{figure}

\section{Introduction}
Industrial anomaly detection (AD)~\cite{19}, as an important task in the computer vision field, plays a crucial role in modern manufacturing and production. The AD task aims to detect and localize anomalies in industrial product images. Since real-world anomalies from different scenarios~\cite{48,49} usually differ in texture, color, shape, size, etc., acquiring and labeling these defects is costly and labor-intensive. Typically, AD models are directly trained on normal samples to identify anomalous samples~\cite{10}. However, this ideal setting seldom exists in real-world situations, e.g., when there is rare or even no relevant training data in a new industry chain, inspecting product defects becomes infeasible by the aforementioned routine. As such, FSAD is proposed to alleviate the annotation costs and novel anomaly detection issues. 

FSAD aims to detect the anomaly region in a query image by additionally utilizing a few normal support images, where the object class of the support/query images is the same. Recently, large vision-language models (LVLMs)~\cite{zhu2023minigpt,su2023pandagpt}, e.g., CLIP~\cite{5}, have demonstrated robust capabilities in recognizing unknown objects~\cite{1,2} and detecting out-of-distribution data~\cite{3}, which paves the way for advancing AD and FSAD tasks~\cite{xie2021scale, xie2021few} and leads to seminal prompt learning based methods~\cite{6,zhou2024cross, wang2022visual}.  
Most of these methods utilize a two-branch network for taking text and query image as input respectively, and the anomaly map is obtained by performing matrix operation among text and image features. For the text branch, a large number of artificial text prompts are usually designed and/or learned, and then aggregated to achieve the normal/abnormal text features (Fig.~\ref{fig1}(a)). Since the above process is heuristic and/or partially learned, it is difficult to encompass all types of anomalies, e.g., some background region is mis-segmented as anomaly ones in Fig. 1(a). In addition, these text-prompt engineering methods involve human intervention and elaborate design, and thus usually cannot meet the automation requirements in real-world industrial scenarios. For the query image branch, existing methods merely rely on simple adapters~\cite{35,36} to extract image features for afterward matrix operation, e.g., by using sliding windows~\cite{6} or kernels of different sizes~\cite{38} for detecting anomaly regions. Although, multi-layer features~\cite{35,37}, e.g., PromptAD~\cite{37}, have been adopted for accurate anomaly estimates. However, the high-order contextual relationships among different vision layers are still not fully leveraged for better inferring query anomaly regions. 

To address the above challenges, we propose a kernel-aware graph prompt learning framework (KAG-prompt) to comprehensively reason the high-order relationships among cross-layer features for FSAD. KAG-prompt consists of a kernel-aware hierarchical graph (KAHG) module and a multi-information fusion (MIF) module, respectively. As in Fig.~\ref{fig1}(b), solely relying on simple manually designed text prompts, KAG-prompt implicitly learns visual prompts by constructing a prompt graph among different layer features for capturing multi-level contextual information. In this manner, KAG-prompt can transfer more relation context from the seen to unseen domains. Specifically, KAHG first uses convolutional kernels of different sizes for extracting visual features of different layers. These kernel-aware features from each layer represent a node in the graph accordingly, and each node w.r.t. visual features of different layers indicate anomalous regions of different sizes. Meanwhile, the relationships between arbitrary pairs of nodes stand for the edges of the graph. Through message passing over this graph to encourage desirable information interactions among different visual layers, KAG-prompt mines richer relationships between cross-layer features in a structured way, thus leading to a more complete understanding of the anomalous region for accurate localization (in Fig.\ref{fig1}(b), the anomaly region is well estimated in the query image). In addition, traditional methods often rely on the maximum value to assess the anomalousness of the prediction map, however, they may overlook other potentially important information. As such, we also propose to utilize the average of the top-k maxima in the prediction map as a new metric, which effectively integrates the information of multiple important anomaly signals in the prediction map, thus improving the accuracy and robustness of anomaly detection. This top-k-induced strategy together with the global class score leads to our multi-information fusion module.

In summary, our contributions are as follows:
\begin{itemize}
    \item We propose kernel-aware graph prompt learning (KAG-prompt), which constructs a kernel-aware hierarchical graph to capture cross-layer multi-level visual relations, thereby enhancing anomaly detection and recognition capabilities. 
    \item We propose a novel image-level scoring method based on multi-level information fusion. By averaging the top-k maximum values from prediction maps and weighting them with scores from global feature-text alignment calculations, our approach effectively integrates critical anomaly signals from prediction maps, distinguishing it from traditional methods that solely rely on the maximum value.
    \item We reveal the significant potential of leveraging contextual information from different layer features for industrial anomaly detection. Diverging from previous methods that rely on complex text prompts, this inspires us to propose a novel FSAD method based on graph prompt learning.
    \item Extensive experiments on MVTecAD~\cite{48} and VisA~\cite{49} datasets demonstrate that our method achieves state-of-the-art FSAD results. We further validate the effectiveness of KAG-prompt through comprehensive ablation studies.
\end{itemize}

\section{Related Work}
\textbf{Anomaly Detection.} AD methods can be classified into four categories: synthesis-based methods, embedding-based methods, reconstruction-based methods, and knowledge distillation-based methods. Synthesis-based methods create anomalous images by introducing noise into the images to train the model. DRAEM~\cite{10} generates anomalies by sampling from an external texture dataset~\cite{11} and producing an anomaly mask using Perlin~\cite{12} noise. Cutpaste~\cite{13} involves cropping parts of an image and pasting them onto different regions of the same image. NSA~\cite{14} employs Poisson image editing to integrate patches from various images seamlessly. Additionally, some methods~\cite{16,15} introduce noise at the feature level to synthesize anomalies.

Embedding-based methods typically utilize a pre-trained model on ImageNet~\cite{17} to encode normal samples into a high-dimensional feature space and then compute distances in this space to detect anomalies. Padim~\cite{18} employs a multivariate Gaussian distribution to estimate the probability distribution of normal samples. Patchcore~\cite{19} stores representative features in the memory bank. CS-Flow~\cite{20} processes multiple features at different scales simultaneously while using normalized flow as the latent space. CFlow~\cite{21}, also based on normalized flow, includes a discriminative encoder and a multi-scale generative decoder. InReaCh~\cite{22} associates patches to channels, considering only channels with high span and low propagation as normal. 

Reconstruction-based methods primarily focus on reconstructing normal inputs, with significant reconstruction errors observed in anomalies. Earlier methods~\cite{40,41,42,43,44} usually employ AE~\cite{23}, VAE~\cite{24} and GAN~\cite{25}. Inspired by the success of diffusion models~\cite{26,27} in generating high-quality and diverse images, more and more methods utilize diffusion models for anomaly modeling. AnoDDPM~\cite{28} is the first to apply diffusion models to the AD task. DiffusionAD~\cite{29} utilizes generated anomaly samples and labels to achieve denoising and segmentation through two sub-networks. DiAD~\cite{30} uses SG networks to reconstruct anomalous regions while preserving the semantic information of the original image.

Knowledge distillation-based methods enable the student network to learn exclusively from the normal samples provided by the teacher network, with anomaly detection achieved by examining the discrepancies between the teacher and student models. RD~\cite{31} employs a reverse distillation paradigm where the input of the student network becomes the embedding of the teacher model, to recover the multi-scale representation of the teacher. RD++~\cite{47} introduces a multi-scale projection layer based on RD and incorporates several loss constraints. MemKD~\cite{32} reinforces the normalcy of students' extracted features by recalling stored normal information.

These aforementioned methods require a substantial quantity of normal samples to accurately model their distribution, rendering them unsuitable for dynamic production environments. In contrast, our FSAD approach uses only a small number of normal samples for inference.

\textbf{Few-Shot Anomaly Detection.} FSAD is first investigated by RegAD~\cite{34}. Patchcore~\cite{19} also demonstrates FSAD performance but with poor results. FastRecon~\cite{46} employs distribution regularization to derive the optimal transformation from support image features to query image features. MuSc~\cite{45} utilizes patches of test images to evaluate each other, thereby constructing a normal distribution. WinCLIP~\cite{6} develops the potential for language-driven anomaly detection by manually designing text prompts for both normal and anomalous cases and utilizing sliding windows to extract and aggregate multi-scale image features. APRIL-GAN~\cite{35} utilizes learnable linear layers to align patch-level image features with textual features, addressing the inefficiencies associated with WinCLIP's multiple windows and further enhancing performance. AnoamlyGPT~\cite{36} proposes a decoder based on visual and textual feature matching to generate pixel-level anomaly localization results. The original image and decoder output serve as inputs to LVLM for anomaly detection, eliminating the need for manual threshold setting. PromptAD~\cite{37} constructs a large number of negative samples by concatenating normal prompts with abnormal suffixes, thereby guiding text prompt learning. Additionally, it introduces the concept of explicit abnormal edges. 

These FSAD methods focus on how to adjust the text prompts to align with the image, ignoring the fact that the images in the downstream task are very different from natural images, and the relations among visual features from different layers are not well leveraged. In contrast, our approach utilizes graph prompt learning to endow desirable informative interactions between different layers, capturing cross-layer multi-level visual relations so that the updated features can be better aligned with texts.

\section{Method}
\subsection{Overview}
\begin{figure*}[t!]
\centering
\includegraphics[width=1\linewidth,height=10.5cm]{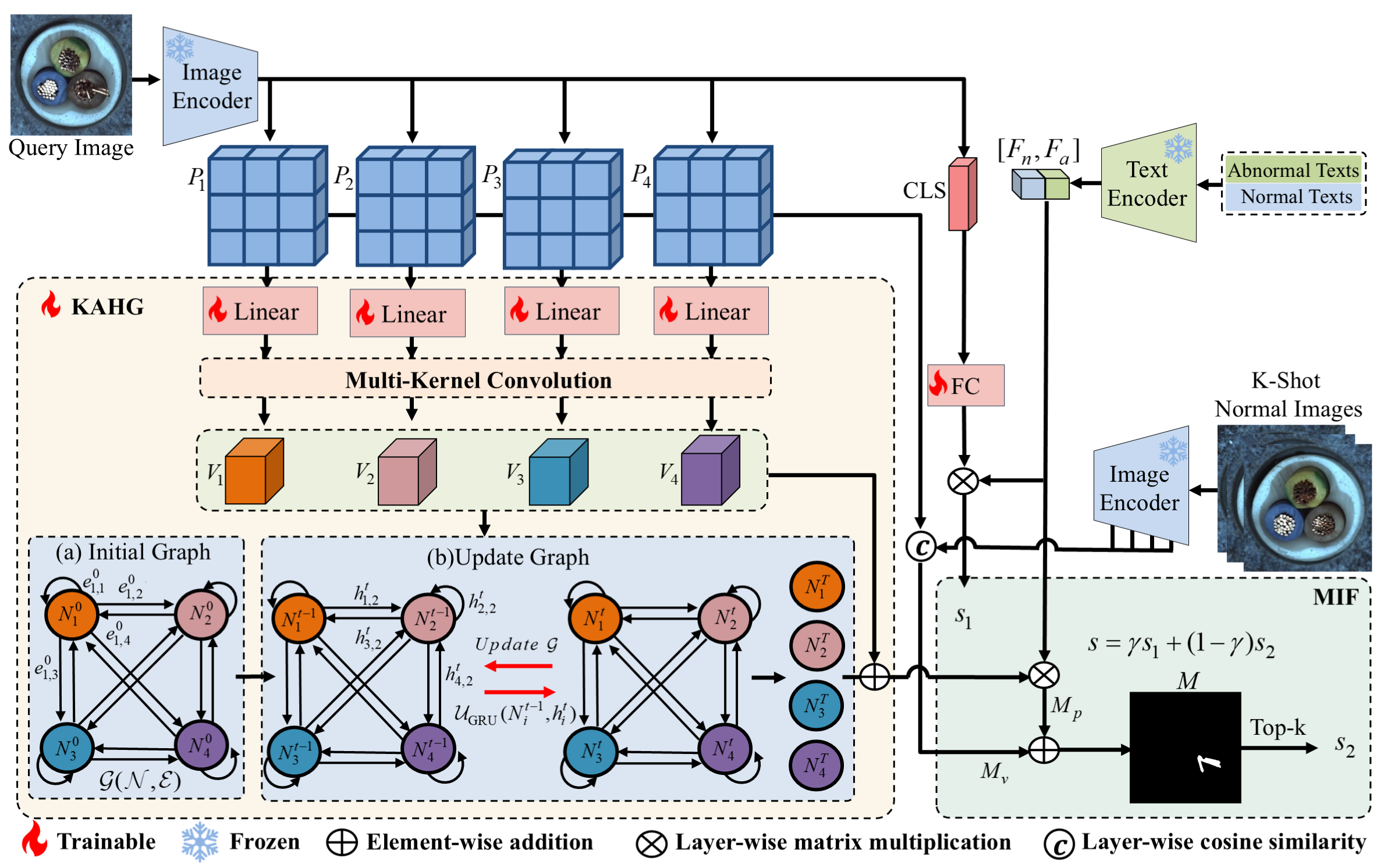}
\caption{\textbf{The architecture of KAG-prompt}. KAG-prompt contains two modules, i.e., KAHG and MIF. The KAHG module takes visual features from different layers as input and these features undergo information interaction within the kernel-aware hierarchical graph $\mathcal{G}=(\mathcal{N, E})$ before aligning with texts to obtain an anomaly localization map $M_p$. Next, the distance between the query image and the most similar patch feature in the memory bank is calculated to get the localization map $M_v$. In the MIF module, for image-level score calculation, the cls token is first adapted and aligned with the texts to get $s_{1}$; then, $M_p$ and $M_v$ are fused to get $s_{2}$ by a top-k fusion mechanism; finally, $s_1$ and $s_2$ are fused to achieve the image-level score $s$.}
\label{fig2}
\end{figure*}

An overview of our proposed KAG-prompt is shown in Fig.~\ref{fig2}. Given a query image $x\in \mathbb{R}^{C\times H\times W}$, it is fed into the image encoder to obtain the patch features $P_{i}, i\in \left \{ 1,2,3,4 \right \} $ in each layer. $P_{i}$ is first passed through the linear layer and then fed to the multi-kernel convolution to obtain $V_{i}$ which focuses on anomalous regions of different sizes. $V_{i}$ is used as an initial node to construct a kernel-aware hierarchical graph, which enables informative interactions among cross-layer visual features through a message passing mechanism to achieve updated visual feature $N_{i}$. Finally, $N_{i}$ is aligned with textual features to obtain pixel-level anomaly localization result $M_{p}$. Additionally, storing patch features of each layer of normal samples in a memory bank, the anomaly localization result $M_{v}$ is obtained by measuring the similarity between query image patches and the most similar patches in the memory bank. The final anomaly localization result $M$ is the fusion of $M_{p}$ and $M_{v}$. For image-level scoring, the cls token is aligned with text features after adapter to get image-level score $s_{1}$, meanwhile, the average of the top-k maximum values of $M$ is taken as $s_{2}$; finally, $s_{1}$ and $s_{2}$ are fused to get the final image-level score.

\subsection{Kernel-Aware Hierarchical Graph}
\textbf{Multi-Kernel Convolution.} Inspired by FiLo~\cite{38}, we also use multi-shape and multi-scale convolutional kernels to focus on anomalous regions of different sizes. Specifically, for a query image $x\in \mathbb{R}^{C\times H\times W}$, which is input to an image encoder, the intermediate patch feature $P_{i}$ is obtained at each stage $i,i\in \left \{ 1,2,3,4 \right \} $. Since the intermediate features are not subjected to final image-text alignment, we first pass $P_{i}$ through the linear layer so that it is in the same dimension as the text, followed by convolutional kernels of different shapes and sizes to obtain $V_{i,k}$. Subsequently, these features are aggregated and normalized to obtain $V_{i}$. At this point, $V_{i}$ not only has the semantic information of the level but also pays attention to the anomalous regions of different shapes and sizes. $V_{i}$ is defined as follows:
\begin{equation}
V_{i}=\textrm{Norm}(\sum_{k=1}^{n} \mathcal{C}_{k}(\textrm{Linear}(P_{i}))),
\label{eq1}	
\end{equation}where the $n$ different convolutional kernels are denoted as $\mathcal{C}_{k}(\cdot)$ and $k$ takes values from $1$ to $n$, and $\textrm{Norm}(\cdot)$ represents the normalization operation.

\noindent\textbf{Kernel-Aware Hierarchical Node Embedding.} We use the patch features of each layer that have gone through the multi-kernel convolution as nodes. For node $N_{i}$, its initialization $N_{i}^{0} $ is denoted as:
\begin{equation}
N_{i}^{0} =V_{i}\in \mathbb{R}^{C'\times H'\times W'},
\label{eq2}	
\end{equation}where $C'$is the channel, and $H'\times W'$ is the spatial resolution.

\noindent\textbf{Edge Embedding.} The kernel-aware hierarchical graph $\mathcal{G}=(\mathcal{N,E})$ is a fully connected graph with self-connections. A loop-edge $e_{i,i}$ is an edge that connects itself. Specifically, we utilize the intra-attention mechanism to compute the response at a given position by focusing on all positions of that node to capture the internal structural relationships of the layer's patch features.
\begin{equation}
\begin{aligned}
    e_{i,i}^{t} &=  \mathcal{F}_\textrm{intra-att}(N_{i}^{t}) \in \mathbb{R}^{C' \times H' \times W'} \\
    &= \alpha \ \textrm{softmax}((W_{c_{1}} * N_{i}^{t})(W_{c_{2}} * N_{i}^{t})^{\mathsf{T}})(W_{c_{3}} * N_{i}^{t}) \\
    &\quad + N_{i}^{t},
\end{aligned}
\label{eq3}
\end{equation}where $\alpha$ is a learnable scale parameter, $W_{c_{j}}$ denotes the learnable convolutional kernel and $*$ denotes the convolution operation.

The line-edge $e_{i,j}$ represents the directed connection from node $N_{i}$ to $N_{j}$ and is used to capture the relationship between them. Specifically, we utilize the inter-attention mechanism to construct the line-edge:
\begin{equation}
    e_{i,j}^{t}=\mathcal{F}_\textrm{inter-att}(N_{i}^{t},N_{j}^{t})=N_{i}^{t} W_{c} {N_{j}^{t}}^{\mathsf{T}}\in \mathbb{R}^{W'H'\times W'H'},
    \label{eq4}
\end{equation}

\begin{equation}
    e_{j,i}^{t}=\mathcal{F}_\textrm{inter-att}(N_{j}^{t},N_{i}^{t})=N_{j}^{t} W_{c}^{\mathsf{T}}{N_{i}^{t}}^{\mathsf{T}}\in \mathbb{R}^{W'H'\times W'H'},
    \label{eq5}
\end{equation}where $W_{c}\in \mathbb{R}^{C' \times C'}$ is the learnable weight matrix. $N_{i}^{t} $ and $N_{j}^{t}$ are flattened into matrices of shape $W'H'\times C'$. By focusing on each pair of nodes, $e_{i,j}^{t}$ reacts to the remote relationships between nodes in different layers.

\noindent\textbf{Hierarchical Node Message Passing.} Since the loop-edge $e_{i,i}$ itself contains the raw and contextual information of the node in that layer, we consider itself as the message $h_{i,i}^{t}$ delivered by the loop-edge:
\begin{equation}
    h_{i,i}^{t}=e_{i,i}^{t-1} \in \mathbb{R}^{C' \times H' \times W'}.
    \label{eq6}
\end{equation}

For  the message passed from node $N_{j}$ to $N_{i}$, we instead assign its edges as weighted weights to neighboring nodes:
\begin{equation}
    \begin{aligned}
    h_{j,i}^{t}&=\mathcal{F}_\textrm{mes}(N_{j}^{t-1},e_{i,j}^{t-1})\\
    &=\mathcal{F}_\textrm{reshape}(\textrm{softmax}(e_{i,j}^{t-1})N_{j}^{t-1}) \in \mathbb{R}^{C' \times H' \times W'},
    \label{eq7}
    \end{aligned}
\end{equation}where $\textrm{softmax}(\cdot )$ is normalized to each row of the input.

Considering that the noise present in layer-based nodes can adversely affect the information in the message passing, we use learnable gating to measure the confidence of the message:
\begin{equation}
    a_{j,i}^{t}=\mathcal{F}_\textrm{gate}(h_{j,i}^{t})=\sigma (\mathcal{F}_\textrm{GAP}(W_\textrm{GAP}*h_{j,i}^{t}+b))\in[0,1],
    \label{eq8}
\end{equation}where $\mathcal{F}_\textrm{GAP}(\cdot)$ is the global average pooling operation, $W_\textrm{GAP}$ and $b$ are its convolutional kernel and bias, and $\sigma$ is the sigmoid activation function. 

Node $N_{i}^{t}$ receives the total messages from neighboring nodes and itself through gating as follows:
\begin{equation}
    h_{i}^{t}=\sum_{j \in \mathcal{N}}a_{j,i}^{t} \odot  h_{j,i}^{t} \in \mathbb{R}^{C' \times H'\times W'},
    \label{eq9}
\end{equation}where $\odot $ is the channel-by-element product.

\noindent\textbf{Hierarchical Nodes Updates.} In iteration $t$, we utilize ConvGRU~\cite{50}, which updates the node state by aggregating the obtained total messages (Eq.~\eqref{eq9}) as well as the node state at step $t-1$:
\begin{equation}
    N_{i}^{t}=\mathcal{U}_\textrm{GRU}(N_{i}^{t-1},h_{i}^{t})\in \mathbb{R}^{C' \times H' \times W'}.
    \label{eq10}
\end{equation}

After the message has been iterated T times, we output the node $N_{i}^{T}$. To preserve the hierarchical and contextual information of the original input, we residually concatenate $N_{i}^{T}$ with the original input $V_{i}$ (Eq.~\eqref{eq1}) to obtain $O_{i}$:
\begin{equation}
    O_{i}=\mathcal{F}_\textrm{reshape}(N_{i}^{T}+V_{i})\in \mathbb{R}^{W'H' \times C'}.
    \label{eq11}
\end{equation}

\subsection{Anomaly Detection By Multi-Information Fusion}
 We use manually designed textual prompts~\cite{6}, processed through a text encoder to obtain normal and abnormal text features, denoted as $F_{text}=[F_{n}, F_{a}]\in \mathbb{R}^{2\times C'}$. We align the layers of visual features with rich contextual information to the text prompts, then aggregate and normalize the predictions $M_{p_{i}}$ from each layer to generate anomaly map $M_{p}$.
\begin{equation}
    M_{p}=\textrm{Up}(\textrm{Norm}\sum_{i=1}^{4}\textrm{softmax}(O_{i}F_{a}^{\mathsf{T}})),
    \label{eq12}
\end{equation}where $\textrm{Up}(\cdot)$ denotes Bilinear interploation.

Additionally, we store the visual features of the support set across layers in a memory bank, denoted as $R$. The localization result $M_{v}$ is obtained by calculating the distance between the query patch and the most similar corresponding patch in $R$.
\begin{equation}
    M_{v}=\textrm{Up}(\sum_{i=1}^{4}(1-\underset{r\in R}{\textrm{max}}(O_{i}r_{i}))).
    \label{eq13}
\end{equation}

The final result of the anomaly localization is:
\begin{equation}
    M=\gamma M_{p}+(1-\gamma )M_{v}.
    \label{eq14}
\end{equation}

For image-level scoring, unlike the previous reliance on maxima to assess the anomalousness of the prediction map, this may overlook other potentially important information. For this reason, we propose utilizing the average of the top-k maximum values in the prediction map as a new metric, the approach that effectively synthesizes information from multiple important anomaly signals in the prediction map. Specifically, the cls feature is aligned to the text features after passing through an adapter (fully connected layer, FC) to obtain a global score $s_{1}$. Next, we average the top-k maxima in $M$ (Eq.~\eqref{eq14}) to obtain $s_{2}$. The final image-level score is the fusion of $s_{1}$ and $s_{2}$:
\begin{equation}
    s_{1}=\textrm{FC}(F_{cls})F_{a}^{\mathsf{T}},
\end{equation}

\begin{equation}
    s=\gamma s_{1}+ (1-\gamma )\textrm{Mean}(\textrm{Top-k}(M)).
\end{equation}

\subsection{Loss Function}
We mainly use cross-entropy loss, focal loss~\cite{51}, and dice loss~\cite{52}. The overall losses are as follows:

\begin{small}
\begin{equation}
    \begin{aligned}
    \mathcal{L}=&\mathcal{L}_\textrm{CE}(s_{1},c)+\lambda_{1}\sum_{i=1}^{4}(\mathcal{L}_\textrm{Focal}([I-M_{p_{i}},M_{p_{i}}],G))\\
    &+\lambda_{2}\sum_{i=1}^{4}(\mathcal{L}_\textrm{Dice}(M_{p_{i}},G)+\mathcal{L}_\textrm{Dice}(I-M_{p_{i}},I-G)) ,
    \end{aligned}
\end{equation}
\end{small}where $\lambda_{1}$ and $\lambda_{2}$ are set to 1.0 in all the experiments, $[\cdot,\cdot]$ denotes concatenation along the channels, $G$ represents ground truth, $c$ denotes true label, and $I$ represents an all-ones matrix.

\begin{table*}[t]
  \centering
    \caption{Performance comparisons of the FSAD methods on the MVTecAD and VisA datasets. Bold indicates the best performance and underlining indicates sub-optimal results. \dag \ indicates our baseline. }
    \setlength{\tabcolsep}{3mm}
    \begin{tabular}{cccccccccc}
    \toprule
    \multirow{2}[4]{*}{Setup} & \multirow{2}[4]{*}{Method} & \multirow{2}[4]{*}{Public} & \multicolumn{2}{c}{MVTecAD} & \multicolumn{2}{c}{VisA} & \multirow{2}[4]{*}{avg} \\
\cmidrule(r){4-5} \cmidrule(r){6-7}  &       &       & AUROC & pAUROC   & AUROC & pAUROC&\\
    \midrule
    \multirow{7}[4]{*}{1-shot} & SPADE~\cite{54} & arXiv2020 & 81.0    & 91.2  & 79.5  & 95.6  & 86.8 \\
          & PatchCore~\cite{19} &  CVPR2022 & 83.4  & 92.0   & 79.9  & 95.4  & 87.7 \\
          & WinCLIP~\cite{6} & CVPR2023 & 93.1  & 95.2   & 83.8  & 96.4  & 92.1 \\
          & APRIL-GAN~\cite{35} & arXiv2023 & 92.0    & 95.1    & \underline{91.2}  & 96.0  & \underline{93.6} \\
          & AnomalyGPT\dag~\cite{36} & AAAI2024 & 94.1  & 95.3 & 87.4  & 96.2  & 93.3 \\
          & PromptAD~\cite{37} & CVPR2024 & \underline{94.6}  & \underline{95.9}  & 86.9  & \underline{96.7}  & 93.5 \\
\cmidrule{2-8}          & \textbf{KAG-prompt (ours)} & -     & \textbf{95.8} & \textbf{96.2}  & \textbf{91.6} & \textbf{97.0}  & \textbf{95.2} \\
    \midrule
    \multirow{7}[4]{*}{2-shot} & SPADE~\cite{54} & arXiv2020 & 82.9  & 92.0  & 80.7 & 96.2 & 88.0 \\
          & PatchCore~\cite{19} & CVPR2022 & 86.3  & 93.3  & 81.6  & 96.1  & 89.3 \\
          & WinCLIP~\cite{6} & CVPR2023 & 94.4  & 96.0  & 84.6  & 96.8   & 93.0 \\
          & APRIL-GAN~\cite{35} & arXiv2023 & 92.4  & 95.0   & \underline{92.2}  & 96.2 & 94.0 \\
          & AnomalyGPT\dag ~\cite{36} & AAAI2024 & 95.5  & 95.6  & 88.6  & 96.4  & 94.0 \\
          & PromptAD~\cite{37} & CVPR2024 & \underline{95.7}  & \underline{96.2} & 88.3  & \underline{97.1}  & \underline{94.3} \\
\cmidrule{2-8}          & \textbf{KAG-prompt (ours)} & -     & \textbf{96.6} & \textbf{96.5}  & \textbf{92.7} & \textbf{97.4} & \textbf{95.8} \\
    \midrule
    \multirow{7}[4]{*}{4-shot} & SPADE~\cite{54} & arXiv2020 & 84.8 & 92.7  & 81.7 & 96.6  & 89.0 \\
          & PatchCore~\cite{19} & CVPR2022 & 88.8  & 94.3  & 85.3  & 96.8  & 91.3 \\
          & WinCLIP~\cite{6} & CVPR2023 & 95.2  & 96.2 & 87.3  & 97.2 & 94.0 \\
          & APRIL-GAN~\cite{35} & arXiv2023 & 92.8  & 95.9  & \underline{92.2}  & 96.2  & 94.3 \\
          & AnomalyGPT\dag ~\cite{36} & AAAI2024 & 96.3  & 96.2  & 90.6  & 96.7 & \underline{95.0} \\
          & PromptAD~\cite{37} & CVPR2024 & \underline{96.6} & \underline{96.5} & 89.1  & \underline{97.4} & 94.9 \\
\cmidrule{2-8}          & \textbf{KAG-prompt (ours)} & -     & \textbf{97.1}  & \textbf{96.7}  & \textbf{93.3} & \textbf{97.7}  & \textbf{96.2} \\
    \bottomrule
    \end{tabular}%
  \label{tb1}
\end{table*}

\begin{table}[t]
  \centering
  \caption{Module ablation at the 1-shot setting of the VisA dataset.}
    \setlength{\tabcolsep}{1mm}
    \begin{tabular}{ccccccc}
    \toprule
    KAHG   & Global $s_{1}$  & Max  & Top-k Fusion & AUROC & pAUROC \\
    \midrule
          &       & $\checkmark$     &       & 89.7  & 96.5\\
    $\checkmark$     &       & $\checkmark$    &       & 91.3  & 97.0  \\
    $\checkmark$    & $\checkmark$     & $\checkmark$     &       & 91.4  & 97.0  \\
    $\checkmark$     & $\checkmark$     &       & $\checkmark$     & 91.6  & 97.0  \\
    \bottomrule
    \end{tabular}
  \label{tb2}
\end{table}%

\section{Experiments}
\subsection{Settings}
\textbf{Datasets.} We mainly conduct experiments on the MVTecAD~\cite{48} and VisA~\cite{49} datasets. The MVTecAD dataset contains 5,354 high-resolution images of 5 textures and 10 objects. The training set contains 3,629 sample images without anomalies. The test set contains 1,725 images including normal and anomalous samples. The VisA dataset has 12 subsets containing 10,821 high-resolution images, of which 9,621 are normal images and 1,200 are anomalous images. As with AnoamlyGPT~\cite{36}, we use the training set of one dataset as well as the synthesized anomalous images for training and perform few-shot testing on the other dataset.

\noindent\textbf{Evaluation Metrics.} We use area under the receiver operating characteristic (AUROC) as an image-level anomaly detection metric. In addition, we use pixel-wise AUROC (pAUROC) to evaluate anomaly localization.

\begin{figure}[t]
    \centering
    \includegraphics[width=8cm,height=4cm]{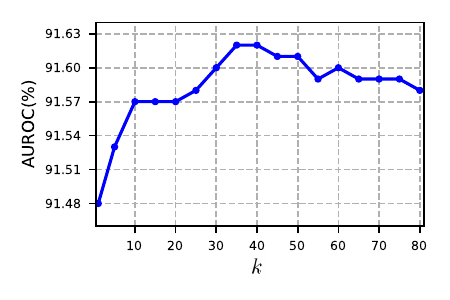}
    \caption{Ablation on top-k strategy $k$ on the 1-shot setting of the VisA dataset.}
    \label{fig7}
\end{figure}

\begin{figure*}[t]
	\centering
	\includegraphics[width=1\linewidth,height=4cm]{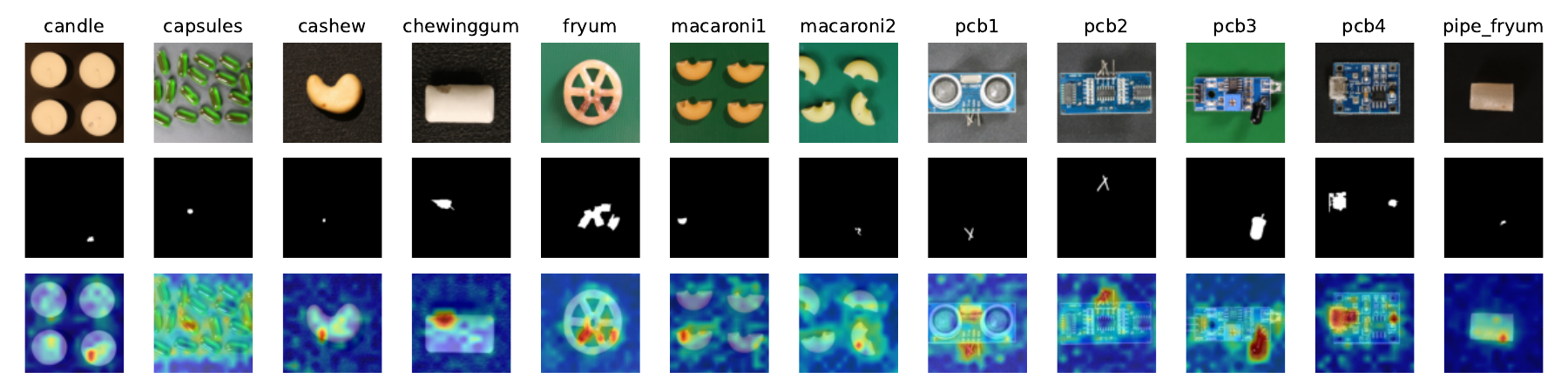}
	\caption{Visualization of KAG-prompt on VisA under 1-shot setting. The first row shows the query image, the second row depicts the corresponding ground truth, and the third row displays the heatmap of abnormal localization by KAG-prompt.}
	\label{fig4}
\end{figure*}

\begin{figure*}[ht!]
    \centering
    \begin{minipage}[b]{0.45\textwidth}
        \centering
        \includegraphics[width=7.5cm,height=4cm]{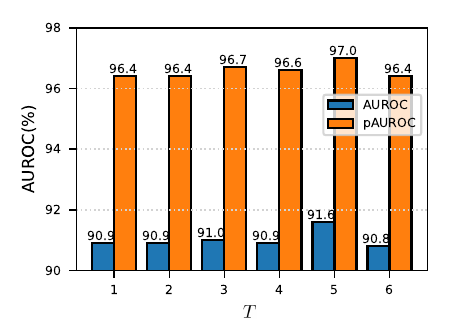}
        \caption{Ablation on graph prompt iterations $T$ at the 1-shot setting on the VisA dataset.}
        \label{fig5}
    \end{minipage}
    \hfill
    \begin{minipage}[b]{0.45\textwidth}
        \centering
        \includegraphics[width=8cm,height=4cm]{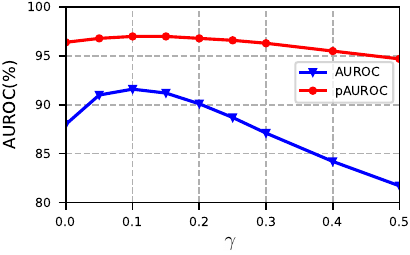}
        \caption{Ablation on the fusion coefficient $\gamma$ at the 1-shot setting on the VisA dataset.}
        \label{fig6}
    \end{minipage}
\end{figure*}

\noindent\textbf{Implementation Details}. Our baseline model is AnomalyGPT~\cite{36}. We synthesize anomaly data for each normal image by NSA~\cite{14} technique and use it to train the model. The image resolution is 224×224. We extract visual features from layers 8, 16, 24, and 36 of the image encoder ImageBind-Huge~\cite{53}. During training, the learning rate is set to 1e-3, batch size to 16, and the number of iterations $T$ for graph prompts is 5. Two RTX-3090 GPUs are used for acceleration during training. The model is trained for 50 epochs on the MVTecAD dataset and 80 epochs on the VisA dataset. We set the fusion coefficient $\gamma$ to 0.1 and select the top-30 scores using a top-k strategy.

\subsection{Comparisons with State-of-the-Arts} Tab.~\ref{tb1} demonstrates the comparison results of KAG-prompt with existing few-shot anomaly detection methods SPADE, PatchCore, Winclip, AnomalyGPT, APRIL-GAN and PromptAD. KAG-prompt shows significant improvements over all methods across both datasets in all metrics. Notably, we achieve a 1.2\% improvement in AUROC and a 0.3\% improvement in pAUROC on MVTecAD with the 1-shot setting compared to the suboptimal method PromptAD. Similarly, KAG-prompt achieves a 4.7\% improvement in AUROC and a 0.3\% improvement in pAUROC compared to PromptAD in the 1-shot setting of VisA. Considering that PromptAD has a higher pixel-level metric but a lower image-level metric on VisA, APRIL-GAN shows better performance in both metrics. Compared to APRIL-GAN, our method improves AUROC by 0.4\% and pAUROC by 1\% on VisA. After averaging the results of all the metrics, our method shows the best results. Compared to the sub-optimal results, KAG-prompt is better by 1.6\%. KAG-prompt similarly reaches the state-of-the-art in 2-/4-shot settings. The quantitative results of KAG-prompt in anomaly localization demonstrate its significant advantages in terms of performance under different numbers of support images.

\subsection{Ablation Study}
\textbf{Module Ablation.} We first verify the effectiveness of different modules of KAG-prompt, including the baseline maximum value calculation (Max), kernel-aware hierarchical graph (KAHG), the cls-guided global scoring $s_{1}$ (Global $s_{1}$), and the top-k strategy for multi-information fusion (Top-k Fusion). The results are shown in Tab.~\ref{tb2}, where each module contributes to the superior performance of KAG-prompt, with kernel-aware hierarchical graph being the most important one. Compared to the baseline, it improves AUROC by 1.6\% and pAUROC by 0.5\%.

\noindent\textbf{Graph Prompt Iterations $T$.} We vary $T$ from 1 to 6 to observe the performance of KAG-prompt. As in Fig.~\ref{fig5}, $T=5$ performs best, thus we set $T=5$ in all our experiments.

\noindent\textbf{Top-k Strategy $k$.} Fig.~\ref{fig7} illustrates the impact of varying values of $k$ on inference performance. We vary $k$ from 1 to 80 and obtain the best result 91.62\% when $k=40$, after which the performance decreases as $k$ increases. As such, we set $k=30$ in all experiments.

\noindent\textbf{The Fusion Coefficient $\gamma$.} We vary $\gamma$ from 0 to 0.5, as shown in Fig.~\ref{fig6}. The results are best at $\gamma=0.1$. Therefore, we set $\gamma=0.1$ for all experiments.

\subsection{Visualization Results}
Fig.~\ref{fig4} shows the visualization results of KAG-prompt on the VisA dataset. KAG-prompt has achieved anomaly localization results that closely match their ground truth. It not only effectively identifies larger anomalous regions but also detects subtle anomalies that are often overlooked. This demonstrates the exceptional anomaly localization capability of KAG-prompt. Such robustness and reliability under few-shot settings highlight the potential practical applicability of KAG-prompt in real-world scenarios. 

\section{Conclusion}
In this paper, we propose a novel anomaly detection method, KAG-prompt. KAG-prompt constructs a kernel-aware hierarchical graph, which learns contextual relationships between cross-layer hierarchical visual features while focusing on anomalous regions of different sizes, to extract updated visual features for aligning them with texts. In addition, vision-guided anomaly detection is introduced to improve the accuracy and robustness of anomaly detection by integrating the information of multiple important anomaly signals through multi-information fusion. Experiments on two commonly used datasets demonstrate the effectiveness of KAG-prompt under FSAD setting.

\bibliography{aaai25}

\begin{thebibliography}{54}
\providecommand{\natexlab}[1]{#1}

\bibitem[{Ballas et~al.(2015)Ballas, Yao, Pal, and Courville}]{50}
Ballas, N.; Yao, L.; Pal, C.; and Courville, A. 2015.
\newblock Delving deeper into convolutional networks for learning video representations.
\newblock \emph{arXiv preprint arXiv:1511.06432}.

\bibitem[{Bergmann et~al.(2019)Bergmann, Fauser, Sattlegger, and Steger}]{48}
Bergmann, P.; Fauser, M.; Sattlegger, D.; and Steger, C. 2019.
\newblock MVTec AD--A comprehensive real-world dataset for unsupervised anomaly detection.
\newblock In \emph{Proceedings of the IEEE/CVF conference on computer vision and pattern recognition}, 9592--9600.

\bibitem[{Chen et~al.(2023)}]{35}
Chen; et~al. 2023.
\newblock A zero-/fewshot anomaly classification and segmentation method for cvpr 2023 vand workshop challenge tracks 1\&2: 1st place on zero-shot ad and 4th place on few-shot ad.
\newblock \emph{arXiv preprint arXiv:2305.17382}, 2(4).

\bibitem[{Cimpoi et~al.(2014)Cimpoi, Maji, Kokkinos, Mohamed, and Vedaldi}]{11}
Cimpoi, M.; Maji, S.; Kokkinos, I.; Mohamed, S.; and Vedaldi, A. 2014.
\newblock Describing textures in the wild.
\newblock In \emph{Proceedings of the IEEE conference on computer vision and pattern recognition}, 3606--3613.

\bibitem[{Cohen et~al.(2020)}]{54}
Cohen; et~al. 2020.
\newblock Sub-image anomaly detection with deep pyramid correspondences.
\newblock \emph{arXiv preprint arXiv:2005.02357}.

\bibitem[{Cohen, Abutbul, and Hoshen(2022)}]{3}
Cohen, N.; Abutbul, R.; and Hoshen, Y. 2022.
\newblock Out-of-distribution detection without class labels.
\newblock In \emph{European Conference on Computer Vision}, 101--117. Springer.

\bibitem[{Creswell et~al.(2018)Creswell, White, Dumoulin, Arulkumaran, Sengupta, and Bharath}]{25}
Creswell, A.; White, T.; Dumoulin, V.; Arulkumaran, K.; Sengupta, B.; and Bharath, A.~A. 2018.
\newblock Generative adversarial networks: An overview.
\newblock \emph{IEEE signal processing magazine}, 35(1): 53--65.

\bibitem[{Defard et~al.(2021)Defard, Setkov, Loesch, and Audigier}]{18}
Defard, T.; Setkov, A.; Loesch, A.; and Audigier, R. 2021.
\newblock Padim: a patch distribution modeling framework for anomaly detection and localization.
\newblock In \emph{International Conference on Pattern Recognition}, 475--489. Springer.

\bibitem[{Dehaene and Eline(2020)}]{40}
Dehaene, D.; and Eline, P. 2020.
\newblock Anomaly localization by modeling perceptual features.
\newblock \emph{arXiv preprint arXiv:2008.05369}.

\bibitem[{Deng and Li(2022)}]{31}
Deng, H.; and Li, X. 2022.
\newblock Anomaly detection via reverse distillation from one-class embedding.
\newblock In \emph{Proceedings of the IEEE/CVF conference on computer vision and pattern recognition}, 9737--9746.

\bibitem[{Deng et~al.(2009)Deng, Dong, Socher, Li, Li, and Fei-Fei}]{17}
Deng, J.; Dong, W.; Socher, R.; Li, L.-J.; Li, K.; and Fei-Fei, L. 2009.
\newblock Imagenet: A large-scale hierarchical image database.
\newblock In \emph{2009 IEEE conference on computer vision and pattern recognition}, 248--255. Ieee.

\bibitem[{Fang et~al.(2023)Fang, Wang, Li, Liu, Hu, and Xiao}]{46}
Fang, Z.; Wang, X.; Li, H.; Liu, J.; Hu, Q.; and Xiao, J. 2023.
\newblock Fastrecon: Few-shot industrial anomaly detection via fast feature reconstruction.
\newblock In \emph{Proceedings of the IEEE/CVF International Conference on Computer Vision}, 17481--17490.

\bibitem[{Girdhar et~al.(2023)Girdhar, El-Nouby, Liu, Singh, Alwala, Joulin, and Misra}]{53}
Girdhar, R.; El-Nouby, A.; Liu, Z.; Singh, M.; Alwala, K.~V.; Joulin, A.; and Misra, I. 2023.
\newblock Imagebind: One embedding space to bind them all.
\newblock In \emph{Proceedings of the IEEE/CVF Conference on Computer Vision and Pattern Recognition}, 15180--15190.

\bibitem[{Gu et~al.(2021)Gu, Lin, Kuo, and Cui}]{1}
Gu, X.; Lin, T.-Y.; Kuo, W.; and Cui, Y. 2021.
\newblock Open-vocabulary object detection via vision and language knowledge distillation.
\newblock \emph{arXiv preprint arXiv:2104.13921}.

\bibitem[{Gu et~al.(2023)Gu, Liu, Chen, Yi, Zhang, Wang, Wang, Shu, Jiang, and Ma}]{32}
Gu, Z.; Liu, L.; Chen, X.; Yi, R.; Zhang, J.; Wang, Y.; Wang, C.; Shu, A.; Jiang, G.; and Ma, L. 2023.
\newblock Remembering normality: Memory-guided knowledge distillation for unsupervised anomaly detection.
\newblock In \emph{Proceedings of the IEEE/CVF International Conference on Computer Vision}, 16401--16409.

\bibitem[{Gu et~al.(2024{\natexlab{a}})Gu, Zhu, Zhu, Chen, Li, Tang, and Wang}]{38}
Gu, Z.; Zhu, B.; Zhu, G.; Chen, Y.; Li, H.; Tang, M.; and Wang, J. 2024{\natexlab{a}}.
\newblock FiLo: Zero-Shot Anomaly Detection by Fine-Grained Description and High-Quality Localization.
\newblock \emph{arXiv preprint arXiv:2404.13671}.

\bibitem[{Gu et~al.(2024{\natexlab{b}})Gu, Zhu, Zhu, Chen, Tang, and Wang}]{36}
Gu, Z.; Zhu, B.; Zhu, G.; Chen, Y.; Tang, M.; and Wang, J. 2024{\natexlab{b}}.
\newblock Anomalygpt: Detecting industrial anomalies using large vision-language models.
\newblock In \emph{Proceedings of the AAAI Conference on Artificial Intelligence}, volume~38, 1932--1940.

\bibitem[{Gudovskiy, Ishizaka, and Kozuka(2022)}]{21}
Gudovskiy, D.; Ishizaka, S.; and Kozuka, K. 2022.
\newblock Cflow-ad: Real-time unsupervised anomaly detection with localization via conditional normalizing flows.
\newblock In \emph{Proceedings of the IEEE/CVF winter conference on applications of computer vision}, 98--107.

\bibitem[{He et~al.(2024)He, Zhang, Chen, Chen, Li, Chen, Wang, Wang, and Xie}]{30}
He, H.; Zhang, J.; Chen, H.; Chen, X.; Li, Z.; Chen, X.; Wang, Y.; Wang, C.; and Xie, L. 2024.
\newblock A diffusion-based framework for multi-class anomaly detection.
\newblock In \emph{Proceedings of the AAAI Conference on Artificial Intelligence}, volume~38, 8472--8480.

\bibitem[{Ho, Jain, and Abbeel(2020)}]{27}
Ho, J.; Jain, A.; and Abbeel, P. 2020.
\newblock Denoising diffusion probabilistic models.
\newblock \emph{Advances in neural information processing systems}, 33: 6840--6851.

\bibitem[{Huang et~al.(2022)Huang, Guan, Jiang, Zhang, Spratling, and Wang}]{34}
Huang, C.; Guan, H.; Jiang, A.; Zhang, Y.; Spratling, M.; and Wang, Y.-F. 2022.
\newblock Registration based few-shot anomaly detection.
\newblock In \emph{European Conference on Computer Vision}, 303--319. Springer.

\bibitem[{Jeong et~al.(2023)Jeong, Zou, Kim, Zhang, Ravichandran, and Dabeer}]{6}
Jeong, J.; Zou, Y.; Kim, T.; Zhang, D.; Ravichandran, A.; and Dabeer, O. 2023.
\newblock Winclip: Zero-/few-shot anomaly classification and segmentation.
\newblock In \emph{Proceedings of the IEEE/CVF Conference on Computer Vision and Pattern Recognition}, 19606--19616.

\bibitem[{Kingma and Welling(2013)}]{24}
Kingma, D.~P.; and Welling, M. 2013.
\newblock Auto-encoding variational bayes.
\newblock \emph{arXiv preprint arXiv:1312.6114}.

\bibitem[{Kuo et~al.(2022)Kuo, Cui, Gu, Piergiovanni, and Angelova}]{2}
Kuo, W.; Cui, Y.; Gu, X.; Piergiovanni, A.; and Angelova, A. 2022.
\newblock F-vlm: Open-vocabulary object detection upon frozen vision and language models.
\newblock \emph{arXiv preprint arXiv:2209.15639}.

\bibitem[{Li et~al.(2021)Li, Sohn, Yoon, and Pfister}]{13}
Li, C.-L.; Sohn, K.; Yoon, J.; and Pfister, T. 2021.
\newblock Cutpaste: Self-supervised learning for anomaly detection and localization.
\newblock In \emph{Proceedings of the IEEE/CVF conference on computer vision and pattern recognition}, 9664--9674.

\bibitem[{Li et~al.(2024{\natexlab{a}})Li, Huang, Xue, and Zhou}]{45}
Li, X.; Huang, Z.; Xue, F.; and Zhou, Y. 2024{\natexlab{a}}.
\newblock Musc: Zero-shot industrial anomaly classification and segmentation with mutual scoring of the unlabeled images.
\newblock \emph{arXiv preprint arXiv:2401.16753}.

\bibitem[{Li et~al.(2024{\natexlab{b}})Li, Zhang, Tan, Chen, Qu, Xie, and Ma}]{37}
Li, X.; Zhang, Z.; Tan, X.; Chen, C.; Qu, Y.; Xie, Y.; and Ma, L. 2024{\natexlab{b}}.
\newblock Promptad: Learning prompts with only normal samples for few-shot anomaly detection.
\newblock In \emph{Proceedings of the IEEE/CVF Conference on Computer Vision and Pattern Recognition}, 16838--16848.

\bibitem[{Liang et~al.(2023)Liang, Zhang, Zhao, Wu, Liu, and Pan}]{43}
Liang, Y.; Zhang, J.; Zhao, S.; Wu, R.; Liu, Y.; and Pan, S. 2023.
\newblock Omni-frequency channel-selection representations for unsupervised anomaly detection.
\newblock \emph{IEEE Transactions on Image Processing}.

\bibitem[{Lin et~al.(2017)Lin, Goyal, Girshick, He, and Doll{\'a}r}]{51}
Lin, T.-Y.; Goyal, P.; Girshick, R.; He, K.; and Doll{\'a}r, P. 2017.
\newblock Focal loss for dense object detection.
\newblock In \emph{Proceedings of the IEEE international conference on computer vision}, 2980--2988.

\bibitem[{Liu et~al.(2022)Liu, Li, Zhao, Du, Jiang, and Geng}]{44}
Liu, T.; Li, B.; Zhao, Z.; Du, X.; Jiang, B.; and Geng, L. 2022.
\newblock Reconstruction from edge image combined with color and gradient difference for industrial surface anomaly detection.
\newblock \emph{arXiv preprint arXiv:2210.14485}.

\bibitem[{Liu et~al.(2023)Liu, Zhou, Xu, and Wang}]{15}
Liu, Z.; Zhou, Y.; Xu, Y.; and Wang, Z. 2023.
\newblock Simplenet: A simple network for image anomaly detection and localization.
\newblock In \emph{Proceedings of the IEEE/CVF Conference on Computer Vision and Pattern Recognition}, 20402--20411.

\bibitem[{McIntosh and Albu(2023)}]{22}
McIntosh, D.; and Albu, A.~B. 2023.
\newblock Inter-realization channels: Unsupervised anomaly detection beyond one-class classification.
\newblock In \emph{Proceedings of the IEEE/CVF International Conference on Computer Vision}, 6285--6295.

\bibitem[{Milletari, Navab, and Ahmadi(2016)}]{52}
Milletari, F.; Navab, N.; and Ahmadi, S.-A. 2016.
\newblock V-net: Fully convolutional neural networks for volumetric medical image segmentation.
\newblock In \emph{2016 fourth international conference on 3D vision (3DV)}, 565--571. Ieee.

\bibitem[{Perlin(1985)}]{12}
Perlin, K. 1985.
\newblock An image synthesizer.
\newblock \emph{ACM Siggraph Computer Graphics}, 19(3): 287--296.

\bibitem[{Radford et~al.(2021)Radford, Kim, Hallacy, Ramesh, Goh, Agarwal, Sastry, Askell, Mishkin, Clark et~al.}]{5}
Radford, A.; Kim, J.~W.; Hallacy, C.; Ramesh, A.; Goh, G.; Agarwal, S.; Sastry, G.; Askell, A.; Mishkin, P.; Clark, J.; et~al. 2021.
\newblock Learning transferable visual models from natural language supervision.
\newblock In \emph{International conference on machine learning}, 8748--8763. PMLR.

\bibitem[{Rombach et~al.(2022)Rombach, Blattmann, Lorenz, Esser, and Ommer}]{26}
Rombach, R.; Blattmann, A.; Lorenz, D.; Esser, P.; and Ommer, B. 2022.
\newblock High-resolution image synthesis with latent diffusion models.
\newblock In \emph{Proceedings of the IEEE/CVF conference on computer vision and pattern recognition}, 10684--10695.

\bibitem[{Roth et~al.(2022)Roth, Pemula, Zepeda, Sch{\"o}lkopf, Brox, and Gehler}]{19}
Roth, K.; Pemula, L.; Zepeda, J.; Sch{\"o}lkopf, B.; Brox, T.; and Gehler, P. 2022.
\newblock Towards total recall in industrial anomaly detection.
\newblock In \emph{Proceedings of the IEEE/CVF conference on computer vision and pattern recognition}, 14318--14328.

\bibitem[{Rudolph, Wandt, and Rosenhahn(2019)}]{23}
Rudolph, M.; Wandt, B.; and Rosenhahn, B. 2019.
\newblock Structuring autoencoders.
\newblock In \emph{Proceedings of the IEEE/CVF International Conference on Computer Vision Workshops}, 0--0.

\bibitem[{Rudolph et~al.(2022)Rudolph, Wehrbein, Rosenhahn, and Wandt}]{20}
Rudolph, M.; Wehrbein, T.; Rosenhahn, B.; and Wandt, B. 2022.
\newblock Fully convolutional cross-scale-flows for image-based defect detection.
\newblock In \emph{Proceedings of the IEEE/CVF Winter Conference on Applications of Computer Vision}, 1088--1097.

\bibitem[{Schlegl et~al.(2017)Schlegl, Seeb{\"o}ck, Waldstein, Schmidt-Erfurth, and Langs}]{42}
Schlegl, T.; Seeb{\"o}ck, P.; Waldstein, S.~M.; Schmidt-Erfurth, U.; and Langs, G. 2017.
\newblock Unsupervised anomaly detection with generative adversarial networks to guide marker discovery.
\newblock In \emph{International conference on information processing in medical imaging}, 146--157. Springer.

\bibitem[{Schl{\"u}ter et~al.(2022)Schl{\"u}ter, Tan, Hou, and Kainz}]{14}
Schl{\"u}ter, H.~M.; Tan, J.; Hou, B.; and Kainz, B. 2022.
\newblock Natural synthetic anomalies for self-supervised anomaly detection and localization.
\newblock In \emph{European Conference on Computer Vision}, 474--489. Springer.

\bibitem[{Su et~al.(2023)Su, Lan, Li, Xu, Wang, and Cai}]{su2023pandagpt}
Su, Y.; Lan, T.; Li, H.; Xu, J.; Wang, Y.; and Cai, D. 2023.
\newblock Pandagpt: One model to instruction-follow them all.
\newblock \emph{arXiv preprint arXiv:2305.16355}.

\bibitem[{Tien et~al.(2023)Tien, Nguyen, Tran, Huy, Duong, Nguyen, and Truong}]{47}
Tien, T.~D.; Nguyen, A.~T.; Tran, N.~H.; Huy, T.~D.; Duong, S.; Nguyen, C. D.~T.; and Truong, S.~Q. 2023.
\newblock Revisiting reverse distillation for anomaly detection.
\newblock In \emph{Proceedings of the IEEE/CVF conference on computer vision and pattern recognition}, 24511--24520.

\bibitem[{Wang et~al.(2020)Wang, Zhang, Guo, and Han}]{41}
Wang, L.; Zhang, D.; Guo, J.; and Han, Y. 2020.
\newblock Image anomaly detection using normal data only by latent space resampling.
\newblock \emph{Applied Sciences}, 10(23): 8660.

\bibitem[{Wang et~al.(2022)Wang, Han, Zhou, and Liu}]{wang2022visual}
Wang, W.; Han, C.; Zhou, T.; and Liu, D. 2022.
\newblock Visual recognition with deep nearest centroids.
\newblock \emph{arXiv preprint arXiv:2209.07383}.

\bibitem[{Wyatt et~al.(2022)Wyatt, Leach, Schmon, and Willcocks}]{28}
Wyatt, J.; Leach, A.; Schmon, S.~M.; and Willcocks, C.~G. 2022.
\newblock Anoddpm: Anomaly detection with denoising diffusion probabilistic models using simplex noise.
\newblock In \emph{Proceedings of the IEEE/CVF Conference on Computer Vision and Pattern Recognition}, 650--656.

\bibitem[{Xie et~al.(2021{\natexlab{a}})Xie, Liu, Xiong, and Shao}]{xie2021scale}
Xie, G.-S.; Liu, J.; Xiong, H.; and Shao, L. 2021{\natexlab{a}}.
\newblock Scale-aware graph neural network for few-shot semantic segmentation.
\newblock In \emph{Proceedings of the IEEE/CVF conference on computer vision and pattern recognition}, 5475--5484.

\bibitem[{Xie et~al.(2021{\natexlab{b}})Xie, Xiong, Liu, Yao, and Shao}]{xie2021few}
Xie, G.-S.; Xiong, H.; Liu, J.; Yao, Y.; and Shao, L. 2021{\natexlab{b}}.
\newblock Few-shot semantic segmentation with cyclic memory network.
\newblock In \emph{Proceedings of the IEEE/CVF International Conference on Computer Vision}, 7293--7302.

\bibitem[{Zavrtanik, Kristan, and Sko{\v{c}}aj(2021)}]{10}
Zavrtanik, V.; Kristan, M.; and Sko{\v{c}}aj, D. 2021.
\newblock Draem-a discriminatively trained reconstruction embedding for surface anomaly detection.
\newblock In \emph{Proceedings of the IEEE/CVF international conference on computer vision}, 8330--8339.

\bibitem[{Zavrtanik, Kristan, and Sko{\v{c}}aj(2022)}]{16}
Zavrtanik, V.; Kristan, M.; and Sko{\v{c}}aj, D. 2022.
\newblock Dsr--a dual subspace re-projection network for surface anomaly detection.
\newblock In \emph{European conference on computer vision}, 539--554. Springer.

\bibitem[{Zhang et~al.(2023)Zhang, Wang, Wu, and Jiang}]{29}
Zhang, H.; Wang, Z.; Wu, Z.; and Jiang, Y.-G. 2023.
\newblock Diffusionad: Denoising diffusion for anomaly detection.
\newblock \emph{arXiv preprint arXiv:2303.08730}, 4.

\bibitem[{Zhou and Wang(2024)}]{zhou2024cross}
Zhou, T.; and Wang, W. 2024.
\newblock Cross-image pixel contrasting for semantic segmentation.
\newblock \emph{IEEE Transactions on Pattern Analysis and Machine Intelligence}.

\bibitem[{Zhu et~al.(2023)Zhu, Chen, Shen, Li, and Elhoseiny}]{zhu2023minigpt}
Zhu, D.; Chen, J.; Shen, X.; Li, X.; and Elhoseiny, M. 2023.
\newblock Minigpt-4: Enhancing vision-language understanding with advanced large language models.
\newblock \emph{arXiv preprint arXiv:2304.10592}.

\bibitem[{Zou et~al.(2022)Zou, Jeong, Pemula, Zhang, and Dabeer}]{49}
Zou, Y.; Jeong, J.; Pemula, L.; Zhang, D.; and Dabeer, O. 2022.
\newblock Spot-the-difference self-supervised pre-training for anomaly detection and segmentation.
\newblock In \emph{European Conference on Computer Vision}, 392--408. Springer.

\end{thebibliography}

\newpage
\begin{strip}
\vspace{-1cm}
\section*{\huge Supplementary Material}
\vspace{2cm}
\end{strip}

\section{Normal and Abnormal Texts}
We generate normal and abnormal texts from a set of prompts. Specifically, the prompts are categorized into (a) state-level and (b) template-level. Normal or abnormal text is generated in the template-level text by replacing the state-level text [s] and the [c] part of the state-level text with the object name. Given that synthetic anomaly samples are used for training and considering the combined effects of graph prompt learning complexity, model training duration, and performance, KAG-prompt employs only the most basic templates, namely, "a photo of a [s]" and "a photo of the [s].". It is experimentally verified that more prompt templates such as "a cropped photo of the [s].", "a close-up photo of a [s]." and "a black and white photo of the [s]." that can further enhance the performance of KAG-prompt.

\begin{figure}[h]
    \centering
    \includegraphics[width=7cm,height=4cm]{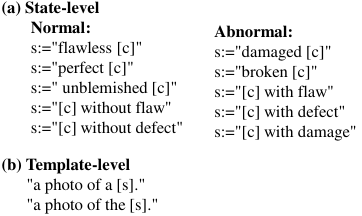}
    \caption{List of normal and abnormal text prompts.}
    \label{fig8}
\end{figure}

\section{Detailed Comparison Results}
In this section, we provide a detailed comparison results of KAG-prompt across subsets of the MVTecAD and VisA datasets. Additionally, we evaluate the performance of KAG-prompt at the image-level precision-recall (AUPR) and pixel-level per-region-overlap (PRO). Specifically, the results for the MVTecAD dataset are presented in \textcolor{red}{Tab.~\ref{mvtec_1shot_auroc}} - \textcolor{red}{\ref{mvtec_4shot_pro}}, while those for the VisA dataset are shown in \textcolor{red}{Tab.~\ref{visa_1shot_auroc}} - \textcolor{red}{\ref{visa_4shot_pro}}. KAG-prompt achieves state-of-the-art (SOTA) performance on nearly all metrics. Although APRIL-GAN performs relatively well on some metrics, it benefits from supervised training on the test set of one dataset and direct testing on another, leveraging real anomaly samples. In contrast, our traditional methods rely solely on normal samples and synthetic anomalies for training. Currently, there is a notable discrepancy between synthetic and real anomalies.

\begin{table}[htbp]
  \centering
  \caption{Ablation on the kernel-aware hierarchical graph at the 1-shot setting of the VisA dataset.}
  \resizebox{\linewidth}{!}{
    \begin{tabular}{cccccc}
    \toprule
    Kernel & Graph & AUROC & pAUROC & PRO   & AUPR \\
    \midrule
          &       & 90.5  & 96.6  & 84.5  & 92.5 \\
          & \checkmark     & 91.5  & 96.8  & 84.6  & 93.2 \\
    \checkmark     & \checkmark     & \textbf{91.6}  & \textbf{97.0}    & \textbf{85.2}  & \textbf{93.2} \\
    \bottomrule
    \end{tabular}
    }
  \label{kernel-graph}
\end{table}

\section{Additional Ablations}
This section presents a comprehensive analysis of the components of the kernel-aware hierarchical graph, with detailed results provided in \textcolor{red}{Tab.~\ref{kernel-graph}}. The hierarchical graph enhances anomaly prediction accuracy by capturing contextual information across layers, leading to improvements of 1\% in AUROC, 0.2\% in pAUROC, 0.1\% in PRO, and 0.7\% in AUPR. Furthermore, incorporating multi-kernel convolution enables visual features at each layer to focus on anomalous regions of different sizes, which improves the segmentation results, thereby enhancing segmentation results with additional improvements of 0.3\% in pAUROC, and 0.6\% in PRO compared to using the graph alone. 

\begin{figure*}
    \centering
    \includegraphics[width=\linewidth]{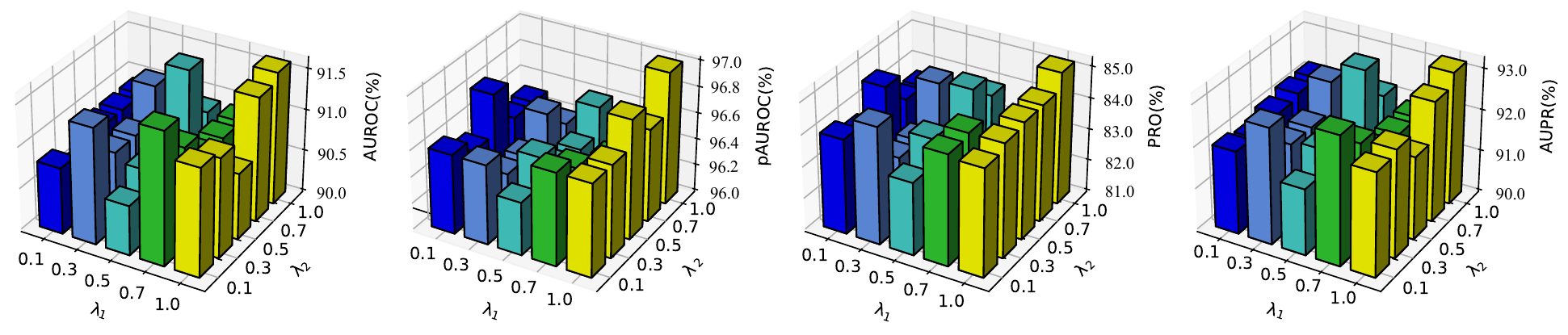}
    \caption{Ablation on the loss coefficients at the 1-shot setting of the VisA dataset.}
    \label{loss}
\end{figure*}

\begin{figure*}[htbp]
    \centering
    \begin{minipage}{0.45\textwidth}
        \centering
        \captionof{table}{Ablation on the learning rate at the 1-shot setting of the VisA dataset.}
        \renewcommand{\arraystretch}{1.2}
        \resizebox{\linewidth}{!}{
            \begin{tabular}{ccccc}
            \toprule
            Learning rate    & AUROC & pAUROC & PRO   & AUPR \\
            \midrule
            1e-1   & 90.7  & 96.7  & 84.1  & 91.9 \\
            1e-2  & 91.1  & 96.5  & 83.6  & 92.3 \\
            1e-3 & \underline{91.6}  & \textbf{97.0}    & \textbf{85.2}  & \underline{93.2} \\
            1e-4  & \textbf{91.9}  & \underline{96.8}  & \underline{85.0}  & \textbf{93.3} \\
            \bottomrule
            \end{tabular}
        }
        \label{lr}
    \end{minipage}\hfill
    \begin{minipage}{0.45\textwidth}
        \includegraphics[width=\linewidth,height=4.5cm]{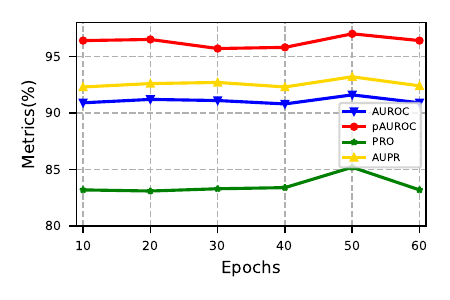}
        \caption{Ablation on the training epochs at the 1-shot setting of the VisA dataset.}
        \label{epochs}
    \end{minipage}
\end{figure*}

In addition, we perform ablation experiments on the learning rate, the training epochs, and the coefficients of the loss function. \textcolor{red}{Tab.~\ref{lr}} demonstrates the effect of the learning rate, considering that the learning rate of 1e-3 yields optimal results, and at the same time pAUROC and PRO perform the best, i.e., the segmentation results are the best, so we set the learning rate to 1e-3 in the experiments. \textcolor{red}{Fig.~\ref{epochs}} demonstrates the effect of the training epochs, and the performance is best when epochs=50. \textcolor{red}{Fig.~\ref{loss}} presents the effect of loss function coefficients, revealing that the best results are obtained when both $\lambda_{1}$ and $\lambda_{2}$ for focal loss and Dice loss are set to 1.

\section{Additional Visualization Results}
\begin{figure*}[]
    \centering
    \includegraphics[width=\linewidth]{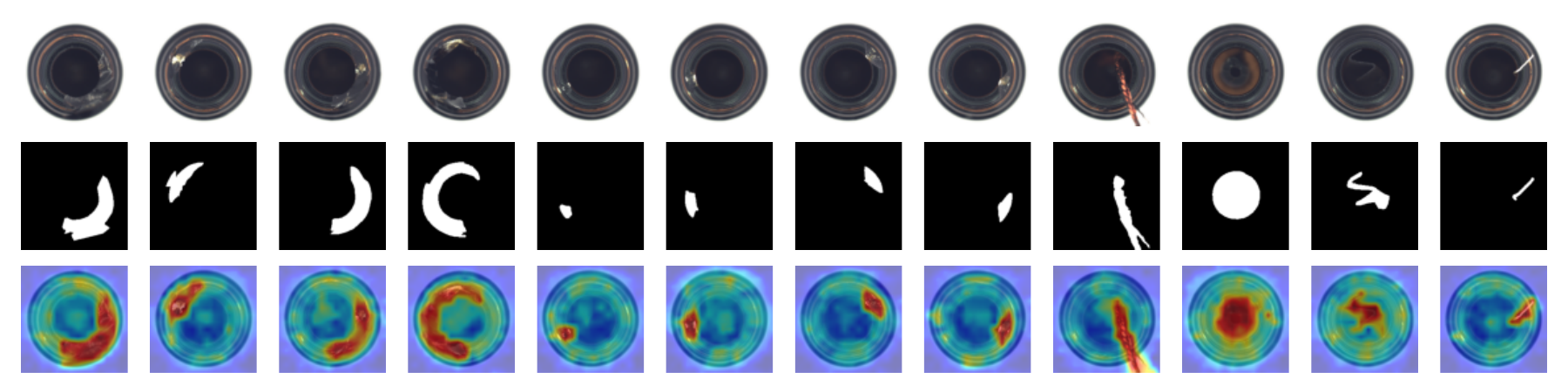}
    \caption{More visualization results for the data subset, bottle, in MVTecAD.}
    \label{mvtec_bottle}
\end{figure*}

\textcolor{red}{Fig.~\ref{mvtec_bottle}} - \textcolor{red}{Fig.~\ref{visa_pipe_fryum}}  further show the visualization results of KAG-prompt for anomaly localization on MVTecAD and VisA datasets. The three rows, from top to bottom, display the test sample, the ground truth, and the heatmap of anomaly localization produced by KAG-prompt, respectively. KAG-prompt can effectively localize not only large defects but also small anomalies. This is attributed to the fact that our approach utilizes the kernel-aware hierarchical graph to focus on anomaly areas of different sizes and combines cross-layer contextual information to extract rich visual features that can be aligned with textual descriptions for accurate localization.


\begin{table*}[]
  \centering
  \caption{Subset-wise performance comparison results of the 1-shot setting for AUROC on MVTecAD.}
    \begin{tabular}{cccccccc}
    \toprule
    MVTecAD & \multicolumn{7}{c}{1-shot} \\
\cmidrule{2-8}    AUROC & SPADE & PatchCore & WinCLIP & APRIL-GAN & AnomalyGPT & PromptAD & KAG-prompt \\
    \midrule
    bottle & 98.7  & 99.4  & 98.2  & 99.9  & 98.1  & 99.8  & 98.7 \\
    cable & 71.2  & 88.8  & 88.9  & 92.8  & 88.8  & 94.2  & 90.7 \\
    capsule & 70.2  & 67.8  & 72.3  & 98.5  & 90.1  & 84.6  & 90.9 \\
    carpet & 98.1  & 95.3  & 99.8  & 99.9  & 99.8  & 100.0   & 100.0 \\
    grid  & 40.0    & 63.6  & 99.5  & 74.7  & 98.4  & 99.8  & 99.1 \\
    hazelnut & 95.8  & 88.3  & 97.5  & 92.0    & 100.0   & 99.8  & 100.0 \\
    leather & 100.0   & 97.3  & 99.9  & 99.0    & 100.0   & 100.0   & 100.0 \\
    metal nut & 71.0    & 73.4  & 98.7  & 84.6  & 98.3  & 99.1  & 99.9 \\
    pill  & 86.5  & 81.9  & 91.2  & 83.4  & 94.6  & 92.6  & 94.8 \\
    screw & 46.7  & 44.4  & 86.4  & 88.5  & 74.7  & 65.0    & 73.1 \\
    tile  & 99.9  & 99.0    & 99.9  & 79.8  & 98.3  & 100.0   & 99.8 \\
    toothbrush & 71.7  & 83.3  & 92.2  & 94.3  & 98.4  & 98.9  & 99.2 \\
    transistor & 77.2  & 78.1  & 83.4  & 94.3  & 77.2  & 94.0    & 93.9 \\
    wood  & 98.8  & 97.8  & 99.9  & 98.9  & 98.1  & 97.9  & 98.8 \\
    zipper & 89.3  & 92.3  & 88.8  & 98.5  & 97.2  & 93.9  & 98.9 \\
    \midrule
    mean  & 81.0    & 83.4  & 93.1  & 92.0    & 94.1  & \underline{94.6}  & \textbf{95.8} \\
    \bottomrule
    \end{tabular}
  \label{mvtec_1shot_auroc}
\end{table*}

\begin{table*}[]
  \centering
  \caption{Subset-wise performance comparison results of the 2-shot setting for AUROC on MVTecAD.}
    \begin{tabular}{cccccccc}
    \toprule
    MVTecAD & \multicolumn{7}{c}{2-shot} \\
\cmidrule{2-8}    AUROC & SPADE & PatchCore & WinCLIP & APRIL-GAN & AnomalyGPT & PromptAD & KAG-prompt \\
    \midrule
    bottle & 99.5  & 99.2  & 99.3  & 99.9  & 99.4  & 100.0   & 99.6 \\
    cable & 76.2  & 91.0    & 88.4  & 94.0    & 90.4  & 99.9  & 91.2 \\
    capsule & 70.9  & 72.8  & 77.3  & 98.7  & 89.8  & 100.0   & 91.1 \\
    carpet & 98.3  & 96.6  & 99.8  & 100.0   & 100.0   & 100.0   & 100.0 \\
    grid  & 41.3  & 67.7  & 99.4  & 74.9  & 99.4  & 98.7  & 99.2 \\
    hazelnut & 96.2  & 93.2  & 98.3  & 92.1  & 100.0   & 99.4  & 100.0 \\
    leather & 100.0   & 97.9  & 99.9  & 99.0    & 100.0   & 91.8  & 100.0 \\
    metal nut & 77.0    & 77.7  & 99.4  & 84.1  & 100.0   & 91.3  & 99.9 \\
    pill  & 84.8  & 82.9  & 92.3  & 83.8  & 95.9  & 100.0   & 94.6 \\
    screw & 46.6  & 49.0    & 86.0    & 90.5  & 83.5  & 98.6  & 82.4 \\
    tile  & 99.9  & 98.5  & 99.9  & 82.0    & 99.5  & 93.6  & 99.9 \\
    toothbrush & 78.6  & 85.9  & 97.5  & 94.8  & 99.2  & 71.0    & 98.9 \\
    transistor & 81.3  & 90.0    & 85.3  & 94.7  & 80.4  & 97.5  & 95 \\
    wood  & 99.2  & 98.3  & 99.9  & 99.0    & 99.1  & 97.4  & 98.4 \\
    zipper & 93.3  & 94.0    & 94.0    & 98.7  & 96.2  & 95.8  & 98.5 \\
    \midrule
    mean  & 82.9  & 86.3  & 94.4  & 92.4  & 95.5  & \underline{95.7}  & \textbf{96.6} \\
    \bottomrule
    \end{tabular}
  \label{mvtec_2shot_auroc}
\end{table*}

\begin{table*}[]
  \centering
  \caption{Subset-wise performance comparison results of the 4-shot setting for AUROC on MVTecAD.}
    \begin{tabular}{cccccccc}
    \toprule
    MVTecAD & \multicolumn{7}{c}{4-shot} \\
\cmidrule{2-8}    AUROC & SPADE & PatchCore & WinCLIP & APRIL-GAN & AnomalyGPT & PromptAD & KAG-prompt \\
    \midrule
    bottle & 99.5  & 99.2  & 99.3  & 99.9  & 99.4  & 100.0   & 99.5 \\
    cable & 83.4  & 91.0    & 90.9  & 94.2  & 91.4  & 98.8  & 91.5 \\
    capsule & 78.9  & 72.8  & 82.3  & 98.8  & 91.1  & 100.0   & 92.9 \\
    carpet & 98.6  & 96.6  & 100.0   & 100.0   & 100.0   & 100.0   & 100.0 \\
    grid  & 44.6  & 67.7  & 99.6  & 76.7  & 99.4  & 98.6  & 99.2 \\
    hazelnut & 98.4  & 93.2  & 98.4  & 93.5  & 100.0   & 99.8  & 100.0 \\
    leather & 100.0   & 97.9  & 100.0   & 99.2  & 100.0   & 95.4  & 100.0 \\
    metal nut & 77.8  & 77.7  & 99.5  & 84.1  & 100.0   & 91.5  & 100.0 \\
    pill  & 86.7  & 82.9  & 92.8  & 84.1  & 95.8  & 99.8  & 95.1 \\
    screw & 50.5  & 49.0    & 87.9  & 91.0    & 84.8  & 100.0   & 84.7 \\
    tile  & 100.0   & 98.5  & 99.9  & 83.7  & 99.2  & 92.9  & 99.9 \\
    toothbrush & 78.8  & 85.9  & 96.7  & 93.2  & 98.1  & 83.6  & 99.2 \\
    transistor & 81.4  & 90.0    & 85.7  & 95.4  & 88.3  & 98.1  & 96.9 \\
    wood  & 98.9  & 98.3  & 99.8  & 99.1  & 99.5  & 95.6  & 98.6 \\
    zipper & 95.1  & 94.0    & 94.5  & 98.7  & 97.5  & 95.0    & 99.1 \\
    \midrule
    mean  & 84.8  & 88.8  & 95.2  & 92.8  & 96.3  & \underline{96.6}  & \textbf{97.1} \\
    \bottomrule
    \end{tabular}
  \label{mvtec_4shot_auroc}
\end{table*}

\begin{table*}[]
  \centering
  \caption{Subset-wise performance comparison results of the 1-shot setting for pAUROC on MVTecAD.}
    \begin{tabular}{cccccccc}
    \toprule
    MVTecAD & \multicolumn{7}{c}{1-shot} \\
\cmidrule{2-8}    pAUROC & SPADE & PatchCore & WinCLIP & APRIL-GAN & AnomalyGPT & PromptAD & KAG-prompt \\
    \midrule
    bottle & 95.3  & 97.9  & 97.5  & 98.7  & 97.6  & 99.6  & 97.5 \\
    cable & 86.4  & 95.5  & 938   & 96.4  & 93.1  & 98.4  & 91.6 \\
    capsule & 96.3  & 95.6  & 94.6  & 97.6  & 92.3  & 99.5  & 97.4 \\
    carpet & 98.2  & 98.4  & 99.4  & 99.5  & 99.1  & 95.9  & 99.4 \\
    grid  & 80.7  & 58.8  & 96.8  & 90.8  & 95.9  & 95.1  & 97.0 \\
    hazelnut & 97.2  & 95.8  & 98.5  & 97.1  & 98.6  & 97.3  & 98.5 \\
    leather & 99.1  & 98.8  & 99.3  & 96.9  & 99.3  & 93.3  & 99.5 \\
    metal nut & 83.8  & 89.3  & 90.0    & 94.8  & 90.6  & 97.3  & 93.7 \\
    pill  & 89.4  & 93.1  & 96.4  & 80.4  & 95.9  & 98.4  & 96.8 \\
    screw & 94.8  & 89.6  & 94.5  & 89.6  & 94.8  & 91.4  & 96.5 \\
    tile  & 91.7  & 94.1  & 96.3  & 98.2  & 96.0    & 92.8  & 97.7 \\
    toothbrush & 94.6  & 97.3  & 97.8  & 98.4  & 98.2  & 94.0    & 98.6 \\
    transistor & 71.4  & 84.9  & 85.0    & 96.1  & 87.6  & 99.1  & 84.6 \\
    wood  & 93.4  & 92.7  & 94.6  & 95.4  & 96.4  & 89.4  & 96.4 \\
    zipper & 94.9  & 97.4  & 93.9  & 96.1  & 93.8  & 96.6  & 97.3 \\
    \midrule
    mean  & 91.2  & 92.0    & 95.2  & 95.1  & 95.3  & \underline{95.9}  & \textbf{96.2} \\
    \bottomrule
    \end{tabular}
  \label{mvtec_1shot_pauroc}
\end{table*}

\begin{table*}[]
  \centering
  \caption{Subset-wise performance comparison results of the 2-shot setting for pAUROC on MVTecAD.}
    \begin{tabular}{cccccccc}
    \toprule
    MVTecAD & \multicolumn{7}{c}{2-shot} \\
\cmidrule{2-8}    pAUROC & SPADE & PatchCore & WinCLIP & APRIL-GAN & AnomalyGPT & PromptAD & KAG-prompt \\
    \midrule
    bottle & 95.7  & 98.1  & 97.7  & 98.7  & 97.7  & 99.5  & 97.7 \\
    cable & 87.4  & 96.4  & 94.3  & 96.9  & 93.2  & 97.6  & 92.4 \\
    capsule & 96.7  & 96.5  & 96.4  & 97.7  & 92.3  & 99.3  & 97.7 \\
    carpet & 98.3  & 98.5  & 99.3  & 99.5  & 99.4  & 96.1  & 99.5 \\
    grid  & 83.5  & 62.6  & 97.7  & 91.2  & 96.5  & 95.5  & 97.1 \\
    hazelnut & 97.6  & 96.3  & 98.7  & 97.1  & 98.7  & 97.6  & 98.5 \\
    leather & 99.1  & 99.0    & 99.3  & 96.9  & 99.3  & 93.2  & 99.5 \\
    metal nut & 85.8  & 94.6  & 91.4  & 95.0    & 91.6  & 97.4  & 94.5 \\
    pill  & 89.9  & 94.2  & 97.0    & 82.5  & 96.9  & 98.5  & 96.9 \\
    screw & 95.6  & 90.0    & 95.2  & 92.1  & 95.6  & 95.1  & 97.3 \\
    tile  & 92.0    & 94.4  & 96.5  & 98.3  & 96.2  & 94.1  & 97.8 \\
    toothbrush & 96.2  & 97.5  & 98.1  & 98.5  & 98.3  & 95.5  & 98.7 \\
    transistor & 72.8  & 89.6  & 88.3  & 96.4  & 88.4  & 99.0    & 86.1 \\
    wood  & 93.8  & 93.2  & 95.3  & 95.8  & 96.4  & 89.1  & 96.3 \\
    zipper & 95.8  & 98.0    & 94.1  & 96.1  & 93.9  & 95.5  & 97.3 \\
    \midrule
    mean  & 92.0    & 93.3  & 96.0    & 95.5  & 95.6  & \underline{96.2}  & \textbf{96.5} \\
    \bottomrule
    \end{tabular}%
  \label{mvtec_2shot_pauroc}%
\end{table*}%

\begin{table*}[]
  \centering
  \caption{Subset-wise performance comparison results of the 4-shot setting for pAUROC on MVTecAD.}
    \begin{tabular}{cccccccc}
    \toprule
    MVTecAD & \multicolumn{7}{c}{4-shot} \\
\cmidrule{2-8}    pAUROC & SPADE & PatchCore & WinCLIP & APRIL-GAN & AnomalyGPT & PromptAD & KAG-prompt \\
    \midrule
    bottle & 96.1  & 98.2  & 97.8  & 98.7  & 98.2  & 99.5  & 98.0 \\
    cable & 88.2  & 97.5  & 94.9  & 97.2  & 94.3  & 98.2  & 92.5 \\
    capsule & 97.0    & 96.8  & 96.2  & 97.7  & 93.4  & 99.3  & 97.9 \\
    carpet & 98.4  & 98.6  & 99.3  & 99.5  & 99.4  & 96.2  & 99.5 \\
    grid  & 87.2  & 69.4  & 98.0    & 91.8  & 97.6  & 95.2  & 97.6 \\
    hazelnut & 97.7  & 97.6  & 98.8  & 97.5  & 98.8  & 97.9  & 98.7 \\
    leather & 99.1  & 99.1  & 99.3  & 97.6  & 99.3  & 939   & 99.5 \\
    metal nut & 87.1  & 95.9  & 92.9  & 95.5  & 93.4  & 97.7  & 94.6 \\
    pill  & 90.7  & 94.8  & 97.1  & 83.7  & 97.2  & 98.5  & 97.0 \\
    screw & 96.4  & 91.3  & 96.0    & 93.1  & 96.8  & 94.9  & 97.5 \\
    tile  & 92.2  & 94.6  & 96.6  & 98.5  & 96.3  & 94.1  & 97.8 \\
    toothbrush & 97.0    & 98.4  & 98.4  & 98.8  & 98.3  & 96.2  & 98.7 \\
    transistor & 73.4  & 90.7  & 88.5  & 96.6  & 90.2  & 99.0    & 86.6 \\
    wood  & 93.9  & 93.5  & 95.4  & 96.0    & 96.4  & 90.6  & 96.2 \\
    zipper & 96.2  & 98.1  & 94.2  & 96.2  & 94.1  & 96.6  & 97.7 \\
    \midrule
    mean  & 92.7  & 94.3  & 96.2  & 95.9  & 96.2  & \underline{96.5}  & \textbf{96.7} \\
    \bottomrule
    \end{tabular}%
  \label{mvtec_4shot_pauroc}%
\end{table*}%

\begin{table*}[]
  \centering
  \caption{Subset-wise performance comparison results of the 1-shot setting for AUPR on MVTecAD.}
    \begin{tabular}{cccccccc}
    \toprule
    MVTecAD & \multicolumn{7}{c}{1-shot} \\
\cmidrule{2-8}    AUPR  & SPADE & PatchCore & WinCLIP & APRIL-GAN & AnomalyGPT & PromptAD & KAG-prompt \\
    \midrule
    bottle & 99.6  & 99.8  & 99.4  & 100.0   & 98.7  & 99.8  & 99.6 \\
    cable & 79.6  & 93.8  & 93.2  & 97.8  & 93.1  & 95.5  & 94.5 \\
    capsule & 91.2  & 89.4  & 91.6  & 99.3  & 97.8  & 97.8  & 98 \\
    carpet & 99.4  & 98.7  & 99.9  & 100.0   & 99.9  & 100.0   & 100.0 \\
    grid  & 66.9  & 81.1  & 99.9  & 84.8  & 98.6  & 98.8  & 99.7 \\
    hazelnut & 97.9  & 92.9  & 98.6  & 98.3  & 100.0   & 99.7  & 100.0 \\
    leather & 100.0   & 99.1  & 100.0   & 99.6  & 100.0   & 100.0   & 100.0 \\
    metal nut & 91.7  & 91.0    & 99.7  & 97.0    & 98.4  & 99.6  & 100.0 \\
    pill  & 97.0    & 96.5  & 98.3  & 76.1  & 98.7  & 98.5  & 99.1 \\
    screw & 71.3  & 71.4  & 94.2  & 97.4  & 88.2  & 78.5  & 89.0 \\
    tile  & 100.0   & 99.6  & 100.0   & 91.1  & 99.4  & 100.0   & 99.9 \\
    toothbrush & 88.3  & 93.5  & 96.7  & 97.8  & 99.1  & 98.5  & 99.7 \\
    transistor & 76.2  & 77.7  & 79.0    & 98.3  & 68.5  & 91.2  & 92.1 \\
    wood  & 99.6  & 99.3  & 100.0   & 99.6  & 98.5  & 99.6  & 99.6 \\
    zipper & 96.9  & 97.2  & 96.8  & 99.5  & 99.0    & 99.0    & 99.7 \\
    \midrule
    mean  & 90.6  & 92.2  & 96.5  & 95.8  & 95.9  & \underline{97.1}  & \textbf{98.1} \\
    \bottomrule
    \end{tabular}
  \label{mvtec_1shot_aupr}%
\end{table*}

\begin{table*}[]
  \centering
  \caption{Subset-wise performance comparison results of the 2-shot setting for AUPR on MVTecAD.}
    \begin{tabular}{cccccccc}
    \toprule
    MVTecAD & \multicolumn{7}{c}{2-shot} \\
\cmidrule{2-8}    AUPR  & SPADE & PatchCore & WinCLIP & APRIL-GAN & AnomalyGPT & PromptAD & KAG-prompt \\
    \midrule
    bottle & 99.8  & 99.8  & 99.8  & 100.0   & 99.8  & 99.9  & 99.9 \\
    cable & 84.5  & 95.1  & 92.9  & 98.2  & 93.9  & 96.9  & 95.0 \\
    capsule & 91.6  & 91.0    & 93.3  & 99.4  & 97.7  & 97.0    & 98.0 \\
    carpet & 99.5  & 99.0    & 99.9  & 100.0   & 100.0   & 100.0   & 100.0 \\
    grid  & 68.5  & 84.1  & 99.8  & 85.1  & 99.8  & 99.9  & 99.7 \\
    hazelnut & 98.0    & 96.0    & 99.1  & 98.4  & 100.0   & 99.8  & 100.0 \\
    leather & 100.0   & 99.3  & 100.0   & 99.6  & 100.0   & 100.0   & 100.0 \\
    metal nut & 93.7  & 92.3  & 99.9  & 96.9  & 100.0   & 100.0   & 100.0 \\
    pill  & 96.5  & 96.6  & 98.6  & 77.5  & 99.3  & 97.8  & 99.0 \\
    screw & 71.0    & 72.9  & 94.1  & 97.8  & 94.3  & 86.7  & 94.0 \\
    tile  & 100.0   & 99.4  & 100.0   & 92.4  & 99.8  & 100.0   & 99.9 \\
    toothbrush & 90.8  & 94.1  & 99.0    & 98.0    & 99.7  & 99.3  & 99.6 \\
    transistor & 81.6  & 89.3  & 80.7  & 98.4  & 69.3  & 92.2  & 93.6 \\
    wood  & 99.7  & 99.5  & 100.0   & 99.6  & 99.7  & 99.7  & 99.5 \\
    zipper & 98.2  & 97.8  & 98.3  & 99.6  & 99.0    & 99.3  & 99.6 \\
    \midrule
    mean  & 91.7  & 93.8  & 97.0    & 96.0    & 96.8  & \underline{97.9}  & \textbf{98.5} \\
    \bottomrule
    \end{tabular}%
  \label{mvtec_2shot_aupr}%
\end{table*}%

\begin{table*}[]
  \centering
  \caption{Subset-wise performance comparison results of the 4-shot setting for AUPR on MVTecAD.}
    \begin{tabular}{cccccccc}
    \toprule
    MVTecAD & \multicolumn{7}{c}{4-shot} \\
\cmidrule{2-8}    AUPR  & SPADE & PatchCore & WinCLIP & APRIL-GAN & AnomalyGPT & PromptAD & KAG-prompt \\
    \midrule
    bottle & 99.9  & 99.8  & 99.8  & 100.0   & 99.8  & 100.0   & 99.9 \\
    cable & 88.8  & 97.1  & 94.4  & 98.3  & 95.4  & 97.4  & 95.3 \\
    capsule & 94.4  & 94.9  & 95.1  & 99.4  & 98.5  & 98.6  & 98.4 \\
    carpet & 99.6  & 98.8  & 100.0   & 100.0   & 100.0   & 100.0   & 100.0 \\
    grid  & 68.8  & 86.4  & 99.9  & 86.5  & 99.7  & 99.7  & 99.7 \\
    hazelnut & 99.1  & 97.0    & 99.1  & 98.7  & 100.0   & 99.9  & 100.0 \\
    leather & 100.0   & 99.6  & 100.0   & 99.7  & 100.0   & 100.0   & 100.0 \\
    metal nut & 94.1  & 97.0    & 99.9  & 96.9  & 100.0   & 99.9  & 100.0 \\
    pill  & 97.0    & 96.9  & 98.6  & 78.1  & 99.3  & 98.5  & 99.1 \\
    screw & 73.3  & 71.8  & 94.9  & 97.9  & 94.5  & 93.8  & 94.7 \\
    tile  & 100.0   & 99.6  & 100.0   & 93.2  & 99.7  & 100.0   & 99.9 \\
    toothbrush & 91.3  & 94.8  & 98.7  & 97.3  & 98.3  & 99.7  & 99.7 \\
    transistor & 80.3  & 84.5  & 80.7  & 98.6  & 83.1  & 92.2  & 96.5 \\
    wood  & 99.7  & 99.5  & 99.9  & 99.6  & 99.0    & 99.5  & 99.5 \\
    zipper & 98.6  & 99.5  & 98.5  & 99.6  & 97.3  & 98.5  & 99.8 \\
    \midrule
    mean  & 92.5  & 99.1  & 97.3  & 96.3  & 97.6  & \underline{98.5}  & \textbf{98.8} \\
    \bottomrule
    \end{tabular}
  \label{mvtec_4shot_aupr}
\end{table*}%

\begin{table*}[]
  \centering
  \caption{Subset-wise performance comparison results of the 1-shot setting for PRO on MVTecAD.}
    \begin{tabular}{cccccccc}
    \toprule
    MVTecAD & \multicolumn{7}{c}{1-shot} \\
\cmidrule{2-8}    PRO   & SPADE & PatchCore & WinCLIP & APRIL-GAN & AnomalyGPT & PromptAD & KAG-prompt \\
    \midrule
    bottle & 91.1  & 93.5  & 91.2  & 96.5  & 94.4  & 93.6  & 93.9 \\
    cable & 63.5  & 84.7  & 72.5  & 92.7  & 83.5  & 87.3  & 80.8 \\
    capsule & 92.7  & 83.9  & 85.6  & 92.1  & 85.5  & 80.1  & 93.6 \\
    carpet & 96.1  & 93.3  & 97.4  & 99.0    & 96.8  & 98.3  & 97.3 \\
    grid  & 67.7  & 21.7  & 90.5  & 81.5  & 90.9  & 94.3  & 92.1 \\
    hazelnut & 94.9  & 88.3  & 93.7  & 95.6  & 95.6  & 92.9  & 94.2 \\
    leather & 98.7  & 95.2  & 98.6  & 90.2  & 97.9  & 98.7  & 98.1 \\
    metal nut & 73.4  & 66.7  & 84.7  & 96.6  & 87.0    & 83.1  & 91.5 \\
    pill  & 92.8  & 89.5  & 93.5  & 64.8  & 94.6  & 90.8  & 96.2 \\
    screw & 85.0    & 68.1  & 82.3  & 88.2  & 85.9  & 78.1  & 88.4 \\
    tile  & 84.2  & 82.5  & 89.4  & 92.6  & 90.6  & 90.7  & 93.5 \\
    toothbrush & 83.5  & 79.0    & 85.3  & 93.4  & 90.8  & 90.1  & 91.7 \\
    transistor & 55.3  & 70.9  & 65.0    & 87.2  & 71.2  & 67.5  & 65.0 \\
    wood  & 92.9  & 87.1  & 91.0    & 92.3  & 93.2  & 92.4  & 93.4 \\
    zipper & 86.8  & 91.2  & 86.0    & 95.9  & 84.8  & 81.0    & 91.7 \\
    \midrule
    mean  & 83.9  & 79.27 & 87.1  & \underline{90.6}  & 89.5  & 87.9  & \textbf{90.8} \\
    \bottomrule
    \end{tabular}%
  \label{mvtec_1shot_pro}%
\end{table*}%

\begin{table*}[]
  \centering
  \caption{Subset-wise performance comparison results of the 2-shot setting for PRO on MVTecAD.}
  
    \begin{tabular}{cccccccc}
    \toprule
    MVTecAD & \multicolumn{7}{c}{2-shot} \\
\cmidrule{2-8}    PRO   & SPADE & PatchCore & WinCLIP & APRIL-GAN & AnomalyGPT & PromptAD & KAG-prompt \\
    \midrule
    bottle & 91.8  & 93.9  & 91.8  & 96.7  & 94.7  & 93.9  & 94.4 \\
    cable & 66.7  & 88.5  & 74.7  & 93.6  & 84.2  & 87.8  & 81.7 \\
    capsule & 93.4  & 86.6  & 90.6  & 92.5  & 85.7  & 79.2  & 94.3 \\
    carpet & 96.2  & 93.7  & 97.3  & 99.0    & 97.3  & 98.2  & 97.5 \\
    grid  & 72.1  & 23.7  & 92.8  & 82.9  & 91.3  & 95.0    & 92.0 \\
    hazelnut & 95.6  & 89.8  & 94.2  & 96.0    & 95.7  & 93.4  & 94.2 \\
    leather & 98.8  & 95.9  & 98.3  & 90.9  & 97.9  & 98.7  & 98 \\
    metal nut & 78.1  & 79.6  & 86.7  & 96.7  & 87.8  & 87.7  & 92.3 \\
    pill  & 93.3  & 91.6  & 94.5  & 66.5  & 95.6  & 90.5  & 96.3 \\
    screw & 87.2  & 69.0    & 84.1  & 90.5  & 86.0    & 74.7  & 88.9 \\
    tile  & 84.6  & 82.5  & 89.6  & 93.1  & 90.7  & 90.9  & 93.6 \\
    toothbrush & 87.4  & 81.0    & 84.7  & 93.8  & 90.9  & 91.6  & 91.9 \\
    transistor & 57.6  & 78.8  & 68.6  & 88.7  & 73.5  & 68.1  & 67.1 \\
    wood  & 93.1  & 86.8  & 91.8  & 92.3  & 93.4  & 91.6  & 92.8 \\
    zipper & 89.0    & 92.8  & 86.4  & 96.0    & 84.9  & 86.4  & 91.5 \\
    \midrule
    mean  & 85.7  & 82.3  & 88.4  & \textbf{91.3} & 90.0    & 88.5  & \underline{91.1} \\
    \bottomrule
    \end{tabular}%
    
  \label{mvtec_2shot_pro}%
\end{table*}%

\begin{table*}[]
  \centering
  \caption{Subset-wise performance comparison results of the 4-shot setting for PRO on MVTecAD.}
  
    \begin{tabular}{cccccccc}
    \toprule
    MVTecAD & \multicolumn{7}{c}{4-shot} \\
\cmidrule{2-8}    PRO   & SPADE & PatchCore & WinCLIP & APRIL-GAN & AnomalyGPT & PromptAD & KAG-prompt \\
    \midrule
    bottle & 92.5  & 94.0    & 91.6  & 96.6  & 95.2  & 94.5  & 94.6 \\
    cable & 69.5  & 91.7  & 77.0    & 94.3  & 85.6  & 88.9  & 82.0 \\
    capsule & 94.1  & 87.8  & 90.1  & 92.6  & 87.1  & 88.7  & 94.9 \\
    carpet & 96.3  & 93.9  & 97.0    & 98.9  & 97.3  & 98.2  & 97.4 \\
    grid  & 78.0    & 30.4  & 93.6  & 84.0    & 93.5  & 93.8  & 93.3 \\
    hazelnut & 95.6  & 92.0    & 94.2  & 96.6  & 96.3  & 95.2  & 94.7 \\
    leather & 98.8  & 96.4  & 98.0    & 92.0    & 97.7  & 98.4  & 97.9 \\
    metal nut & 81.2  & 83.8  & 89.4  & 96.7  & 90.1  & 87.6  & 92.4 \\
    pill  & 93.9  & 92.5  & 94.6  & 67.7  & 96.1  & 92.0    & 96.5 \\
    screw & 89.5  & 72.4  & 86.3  & 91.7  & 87.6  & 86.7  & 89.8 \\
    tile  & 84.9  & 83.0    & 89.9  & 93.7  & 90.4  & 90.9  & 93.2 \\
    toothbrush & 89.0    & 85.5  & 86.0    & 94.8  & 90.9  & 91.3  & 91.9 \\
    transistor & 58.5  & 79.5  & 69.0    & 89.2  & 74.9  & 73.0    & 67.4 \\
    wood  & 93.2  & 87.7  & 91.7  & 92.4  & 92.6  & 91.4  & 92.7 \\
    zipper & 90.1  & 93.4  & 86.9  & 96.1  & 85.3  & 87.5  & 92.7 \\
    \midrule
    mean  & 87.0    & 84.3  & 89.0    & \textbf{91.8} & 90.7  & 90.5  & \underline{91.4} \\
    \bottomrule
    \end{tabular}%
    
  \label{mvtec_4shot_pro}%
\end{table*}%

\begin{table*}[]
  \centering
  \caption{Subset-wise performance comparison results of the 1-shot setting for AUROC on VisA.}
    \begin{tabular}{cccccccc}
    \toprule
    VisA  & \multicolumn{7}{c}{1-shot} \\
\cmidrule{2-8}    AUROC & SPADE & PatchCore & WinCLIP & APRIL-GAN & AnomalyGPT & PromptAD & KAG-prompt \\
    \midrule
    candle & 86.1  & 85.1  & 93.4  & 90.9  & 85.75 & 90.3  & 96.3 \\
    capsules & 73.3  & 60.0    & 85.0    & 92.7  & 85.8 & 84.5  & 89.1 \\
    cashew & 95.9  & 89.5  & 94.0    & 93.4  & 91.5 & 95.6  & 94.4 \\
    chewinggum & 92.1  & 97.3  & 97.6  & 97.3  & 98.0    & 96.4  & 98.7 \\
    fryum & 81.1  & 75.0    & 88.5  & 93.4  & 92.2 & 90.3  & 93.7 \\
    macaroni1 & 66.0    & 68.0    & 82.9  & 89.7  & 89.9  & 88.6  & 96.6 \\
    macaroni2 & 55.8  & 55.6  & 70.2  & 79.2  & 84.7 & 69.1  & 84.1 \\
    pcb1  & 87.2  & 78.9  & 75.6  & 89.0    & 82.6 & 88.7  & 84.4 \\
    pcb2  & 73.5  & 81.5  & 62.2  & 85.4  & 75.6 & 71.6  & 83.3 \\
    pcb3  & 72.2  & 82.7  & 74.1  & 87.8  & 75.0 & 97.1  & 81.4 \\
    pcb4  & 93.4  & 93.9  & 85.2  & 97.5  & 88.8 & 91.4  & 97.9 \\
    pipe fryum & 77.9  & 90.7  & 97.2  & 98.3  & 99.3  & 96.9  & 99.5 \\
    \midrule
    mean  & 79.5  & 79.9  & 83.8  & \underline{91.2}  & 87.4  & 86.9  & \textbf{91.6} \\
    \bottomrule
    \end{tabular}%
  \label{visa_1shot_auroc}
\end{table*}

\begin{table*}[]
  \centering
  \caption{Subset-wise performance comparison results of the 2-shot setting for AUROC on VisA.}
    \begin{tabular}{cccccccc}
    \toprule
    VisA  & \multicolumn{7}{c}{2-shot} \\
\cmidrule{2-8}    AUROC & SPADE & PatchCore & WinCLIP & APRIL-GAN & AnomalyGPT & PromptAD & KAG-prompt \\
    \midrule
    candle & 91.3  & 85.3  & 94.8  & 91.3  & 83.1  & 91.0    & 94.5 \\
    capsules & 71.7  & 57.8  & 84.9  & 93.4  & 88.8  & 84.9  & 89.9 \\
    cashew & 97.3  & 93.6  & 94.3  & 93.4  & 93.2  & 94.7  & 94.3 \\
    chewinggum & 93.4  & 97.8  & 97.3  & 97.1  & 98.4  & 96.6  & 98.8 \\
    fryum & 90.5  & 83.4  & 90.5  & 93.1  & 94.1  & 89.2  & 95.1 \\
    macaroni1 & 69.1  & 75.6  & 83.3  & 90.0    & 91.4  & 84.2  & 95.9 \\
    macaroni2 & 58.3  & 57.3  & 71.8  & 81.0    & 85.8  & 82.6  & 86.9 \\
    pcb1  & 86.7  & 71.5  & 76.7  & 90.9  & 83.6  & 90.9  & 88.8 \\
    pcb2  & 70.3  & 84.3  & 62.6  & 89.2  & 78.3  & 73.0    & 82.0 \\
    pcb3  & 75.8  & 84.8  & 78.8  & 90.9  & 79.2  & 76.2  & 89.0 \\
    pcb4  & 86.1  & 94.3  & 82.3  & 97.9  & 88.4  & 97.5  & 97.2 \\
    pipe fryum & 78.1  & 93.5  & 98.0    & 98.4  & 99.4  & 98.9  & 99.7 \\
    \midrule
    mean  & 80.7  & 81.6  & 84.6  & \underline{92.2}  & 88.6  & 88.3  & \textbf{92.7} \\
    \bottomrule
    \end{tabular}
  \label{visa_2shot_auroc}
\end{table*}

\begin{table*}[]
  \centering
  \caption{Subset-wise performance comparison results of the 4-shot setting for AUROC on VisA.}
    \begin{tabular}{cccccccc}
    \toprule
    VisA  & \multicolumn{7}{c}{4-shot} \\
\cmidrule{2-8}    AUROC & SPADE & PatchCore & WinCLIP & APRIL-GAN & AnomalyGPT & PromptAD & KAG-prompt \\
    \midrule
    candle & 92.8  & 87.8  & 95.1  & 91.4  & 85.6  & 93.0    & 95.0 \\
    capsules & 73.4  & 63.4  & 86.8  & 93.7  & 89.0    & 80.6  & 90.1 \\
    cashew & 96.4  & 93.0    & 95.2  & 94.3  & 94.2  & 93.6  & 97.3 \\
    chewinggum & 93.5  & 98.3  & 97.7  & 97.2  & 98.7  & 96.8  & 98.4 \\
    fryum & 92.9  & 88.6  & 90.8  & 93.5  & 95.2  & 89.0    & 94.3 \\
    macaroni1 & 65.8  & 82.9  & 85.2  & 90.1  & 92.1  & 88.2  & 97.6 \\
    macaroni2 & 56.7  & 61.7  & 70.9  & 82.5  & 87.2  & 81.2  & 87.2 \\
    pcb1  & 83.4  & 84.7  & 88.3  & 91.2  & 83.1  & 90.9  & 89.7 \\
    pcb2  & 71.7  & 84.3  & 67.5  & 89.2  & 85.5  & 78.6  & 87.8 \\
    pcb3  & 79.0    & 87.0    & 83.3  & 91.6  & 86.9  & 80.3  & 85.8 \\
    pcb4  & 95.4  & 95.6  & 87.6  & 98.2  & 91.2  & 97.8  & 97.4 \\
    pipe fryum & 79.3  & 96.4  & 98.5  & 98.5  & 99    & 98.6  & 99.5 \\
    \midrule
    mean  & 81.7  & 85.3  & 87.3  & \underline{92.6}  & 90.6  & 89.1  & \textbf{93.3} \\
    \bottomrule
    \end{tabular}
  \label{visa_4shot_auroc}
\end{table*}

\begin{table*}[]
  \centering
  \caption{Subset-wise performance comparison results of the 1-shot setting for pAUROC on VisA.}
    \begin{tabular}{cccccccc}
    \toprule
    VisA  & \multicolumn{7}{c}{1-shot} \\
\cmidrule{2-8}    pAUROC & SPADE & PatchCore & WinCLIP & APRIL-GAN & AnomalyGPT & PromptAD & KAG-prompt \\
    \midrule
    candle & 97.9  & 97.2  & 97.4  & 98.7  & 98.2  & 95.8  & 98.6 \\
    capsules & 95.5  & 93.2  & 96.4  & 98.0    & 97.6  & 95.4  & 97.5 \\
    cashew & 95.9  & 98.1  & 98.5  & 90.8  & 96.5  & 99.1  & 96.8 \\
    chewinggum & 96.0    & 96.9  & 98.6  & 99.7  & 99.1  & 99.1  & 99.2 \\
    fryum & 93.5  & 93.3  & 96.4  & 93.6  & 92.7  & 95.4  & 93.8 \\
    macaroni1 & 97.9  & 95.2  & 96.4  & 99.3  & 97.6  & 97.8  & 98.6 \\
    macaroni2 & 94.1  & 89.1  & 96.8  & 98.4  & 95.3  & 96.6  & 97.3 \\
    pcb1  & 94.7  & 96.1  & 96.6  & 95.3  & 97.9  & 96.6  & 97.4 \\
    pcb2  & 95.1  & 95.4  & 93.0    & 92.5  & 92.5  & 93.5  & 94.0 \\
    pcb3  & 96.0    & 96.2  & 94.3  & 93.2  & 94.6  & 95.9  & 95.6 \\
    pcb4  & 92.0    & 95.6  & 94.0    & 95.8  & 94.9  & 95.5  & 96.9 \\
    pipe fryum & 98.4  & 98.8  & 98.3  & 97.3  & 98.0    & 99.1  & 98.4 \\
    \midrule
    mean  & 95.6  & 95.4  & 96.4  & 96.0    & 96.2  & \underline{96.7}  & \textbf{97.0} \\
    \bottomrule
    \end{tabular}
  \label{visa_1shot_pauroc}
\end{table*}

\begin{table*}[]
  \centering
  \caption{Subset-wise performance comparison results of the 2-shot setting for pAUROC on VisA.}
    \begin{tabular}{cccccccc}
    \toprule
    VisA  & \multicolumn{7}{c}{2-shot} \\
\cmidrule{2-8}    pAUROC & SPADE & PatchCore & WinCLIP & APRIL-GAN & AnomalyGPT & PromptAD & KAG-prompt \\
    \midrule
    candle & 98.1  & 97.7  & 97.7  & 98.6  & 98.3  & 95.9  & 98.8 \\
    capsules & 96.5  & 94.0    & 96.8  & 98.0    & 98.3  & 96.1  & 98.4 \\
    cashew & 95.9  & 98.2  & 98.5  & 90.8  & 95.7  & 99.2  & 98.2 \\
    chewinggum & 96.0    & 96.6  & 98.6  & 99.7  & 98.7  & 99.2  & 99.2 \\
    fryum & 93.9  & 94.0    & 97.0    & 93.6  & 93.6  & 96.4  & 93.8 \\
    macaroni1 & 98.5  & 96.0    & 96.5  & 99.3  & 96.8  & 98.3  & 98.9 \\
    macaroni2 & 95.2  & 90.2  & 96.8  & 98.4  & 96.3  & 97.2  & 97.4 \\
    pcb1  & 96.5  & 97.6  & 97.0    & 95.8  & 96.5  & 96.9  & 98.0 \\
    pcb2  & 95.7  & 96.0    & 93.9  & 92.8  & 93.0    & 94.8  & 95.2 \\
    pcb3  & 96.6  & 97.1  & 95.1  & 93.5  & 95.9  & 96.1  & 95.6 \\
    pcb4  & 92.8  & 96.2  & 95.6  & 95.9  & 95.3  & 95.6  & 97.3 \\
    pipe fryum & 98.7  & 99.1  & 98.5  & 97.3  & 98.2  & 99.4  & 98.4 \\
    \midrule
    mean  & 96.2  & 96.1  & 96.8  & 96.2  & 96.4  & \underline{97.1}  & \textbf{97.4} \\
    \bottomrule
    \end{tabular}%
  \label{visa_2shot_pauroc}%
\end{table*}

\begin{table*}[]
  \centering
  \caption{Subset-wise performance comparison results of the 4-shot setting for pAUROC on VisA.}
    \begin{tabular}{cccccccc}
    \toprule
    VisA  & \multicolumn{7}{c}{4-shot} \\
\cmidrule{2-8}    pAUROC & SPADE & PatchCore & \multicolumn{1}{l}{WinCLIP} & APRIL-GAN & AnomalyGPT & PromptAD & KAG-prompt \\
    \midrule
    candle & 98.2  & 97.9  & 97.8  & 98.7  & 98.6  & 96.0    & 98.8 \\
    capsules & 97.7  & 94.8  & 97.1  & 98.1  & 98.2  & 96.8  & 98.5 \\
    cashew & 95.9  & 98.3  & 98.7  & 90.8  & 96.9  & 99.2  & 98.3 \\
    chewinggum & 95.7  & 96.8  & 98.5  & 99.7  & 98.4  & 99.2  & 99.2 \\
    fryum & 94.4  & 94.2  & 97.1  & 93.7  & 94.3  & 96.6  & 94.7 \\
    macaroni1 & 98.8  & 97.0    & 97.0    & 99.3  & 97.7  & 98.2  & 99.1 \\
    macaroni2 & 96.4  & 93.9  & 97.3  & 98.4  & 96.6  & 97.0    & 97.4 \\
    pcb1  & 96.8  & 98.1  & 98.1  & 96.0    & 96.8  & 98.2  & 97.9 \\
    pcb2  & 96.3  & 96.6  & 94.6  & 93.0    & 93.2  & 95.3  & 96.1 \\
    pcb3  & 96.9  & 97.4  & 95.8  & 93.7  & 96.1  & 96.8  & 96.4 \\
    pcb4  & 94.1  & 97.0    & 96.1  & 96.0    & 95.6  & 96.2  & 97.6 \\
    pipe fryum & 98.8  & 99.1  & 98.7  & 97.4  & 98.5  & 99.3  & 98.5 \\
    \midrule
    mean  & 96.6  & 96.8  & 97.2  & 96.2  & 96.7  & \underline{97.4}  & \textbf{97.7} \\
    \bottomrule
    \end{tabular}%
  \label{visa_4shot_pauroc}%
\end{table*}%

\begin{table*}[]
  \centering
  \caption{Subset-wise performance comparison results of the 1-shot setting for AUPR on VisA.}
    \begin{tabular}{cccccccc}
    \toprule
    VisA  & \multicolumn{7}{c}{1-shot} \\
\cmidrule{2-8}    AUPR  & SPADE & PatchCore & WinCLIP & APRIL-GAN & AnomalyGPT & PromptAD & KAG-prompt \\
    \midrule
    candle & 86.5  & 86.6  & 93.6  & 91.6  & 86.4  & 93.7  & 96.8 \\
    capsules & 79.4  & 72.3  & 89.9  & 95.8  & 91.4  & 90.1  & 94.1 \\
    cashew & 97.9  & 94.6  & 97.2  & 97.1  & 96.3  & 97.6  & 97.6 \\
    chewinggum & 96.4  & 98.9  & 99.0    & 99.0    & 99.1  & 99.1  & 99.4 \\
    fryum & 89.8  & 87.6  & 94.7  & 97.3  & 96.4  & 93.8  & 97.5 \\
    macaroni1 & 61.9  & 67.8  & 84.9  & 92.0    & 92.8  & 86.3  & 97.0 \\
    macaroni2 & 52.7  & 54.9  & 68.4  & 83.4  & 87.1  & 72.5  & 88.1 \\
    pcb1  & 84.9  & 72.1  & 76.5  & 90.2  & 76.0    & 88.0    & 81.0 \\
    pcb2  & 74.9  & 84.4  & 64.9  & 87.1  & 76.9  & 75.4  & 86.8 \\
    pcb3  & 75.5  & 84.6  & 73.5  & 89.4  & 79.9  & 75.2  & 83.2 \\
    pcb4  & 92.9  & 92.8  & 78.5  & 97.0    & 81.9  & 90.5  & 97.5 \\
    pipe fryum & 88.3  & 95.4  & 98.6  & 99.2  & 99.7  & 98.3  & 99.8 \\
    \midrule
    mean  & 82.0    & 82.8  & 85.1  & \textbf{93.3} & 88.7  & 88.4  & \underline{93.2} \\
    \bottomrule
    \end{tabular}%
  \label{visa_1shot_aupr}%
\end{table*}%

\begin{table*}[]
  \centering
  \caption{Subset-wise performance comparison results of the 2-shot setting for AUPR on VisA.}
    \begin{tabular}{cccccccc}
    \toprule
    VisA  & \multicolumn{7}{c}{2-shot} \\
\cmidrule{2-8}    AUPR  & SPADE & PatchCore & WinCLIP & APRIL-GAN & AnomalyGPT & PromptAD & KAG-prompt \\
    \midrule
    candle & 90.7  & 86.8  & 95.1  & 91.9  & 83.2  & 93.6  & 95.3 \\
    capsules & 79.9  & 73.6  & 88.9  & 96.3  & 94.3  & 88.3  & 95.0 \\
    cashew & 98.6  & 96.9  & 97.3  & 97.2  & 96.1  & 97.4  & 97.3 \\
    chewinggum & 97.1  & 99.1  & 98.9  & 98.9  & 99.1  & 98.4  & 99.5 \\
    fryum & 94.5  & 92.1  & 95.8  & 97.2  & 97.5  & 96.0    & 97.9 \\
    macaroni1 & 64.5  & 74.9  & 84.7  & 92.2  & 92.8  & 91.1  & 97.0 \\
    macaroni2 & 55.9  & 57.2  & 70.4  & 85.1  & 88.6  & 84.7  & 89.8 \\
    pcb1  & 83.8  & 72.6  & 78.3  & 91.6  & 78.7  & 80.9  & 86.7 \\
    pcb2  & 71.7  & 86.6  & 65.8  & 91.3  & 75.8  & 73.0    & 85.2 \\
    pcb3  & 78.3  & 86.1  & 80.9  & 92.2  & 80.4  & 82.8  & 90.0 \\
    pcb4  & 81.9  & 93.2  & 72.5  & 97.8  & 81.8  & 94.5  & 96.8 \\
    pipe fryum & 88.1  & 96.8  & 99.0    & 99.3  & 99.7  & 99.1  & 99.9 \\
    \midrule
    mean  & 82.3  & 84.8  & 85.8  & \textbf{94.2} & 89.0    & \underline{90.0}    & \textbf{94.2} \\
    \bottomrule
    \end{tabular}
  \label{visa_2shot_aupr}
\end{table*}

\begin{table*}[]
  \centering
  \caption{Subset-wise performance comparison results of the 4-shot setting for AUPR on VisA.}
    \begin{tabular}{cccccccc}
    \toprule
    VisA  & \multicolumn{7}{c}{4-shot} \\
\cmidrule{2-8}    AUPR  & SPADE & PatchCore & WinCLIP & APRIL-GAN & AnomalyGPT & PromptAD & KAG-prompt \\
    \midrule
    candle & 92.6  & 88.9  & 95.3  & 92.0    & 86.4  & 92.9  & 95.7 \\
    capsules & 81.8  & 78.4  & 91.5  & 96.5  & 94.3  & 89.8  & 95.0 \\
    cashew & 98.3  & 96.5  & 97.7  & 97.6  & 97.3  & 97.0    & 98.6 \\
    chewinggum & 97.1  & 99.3  & 99.0    & 99.0    & 99.2  & 98.5  & 99.3 \\
    fryum & 95.8  & 95.0    & 96.0    & 97.4  & 97.9  & 93.6  & 97.8 \\
    macaroni1 & 60.2  & 82.1  & 86.5  & 92.4  & 92.7  & 98.2  & 98.0 \\
    macaroni2 & 51.9  & 60.2  & 69.6  & 86.3  & 89.0    & 82.2  & 89.2 \\
    pcb1  & 83.2  & 81.0    & 87.7  & 91.8  & 77.1  & 90.1  & 87.6 \\
    pcb2  & 74.2  & 86.2  & 71.3  & 91.3  & 86.7  & 75.3  & 90.7 \\
    pcb3  & 81.0    & 88.3  & 84.8  & 92.7  & 88.3  & 83.5  & 86.9 \\
    pcb4  & 94.8  & 94.9  & 85.6  & 98.0    & 87.1  & 97.5  & 97.0 \\
    pipe fryum & 88.8  & 98.3  & 99.2  & 99.4  & 99.5  & 99.3  & 99.8 \\
    \midrule
    mean  & 83.4  & 87.5  & 88.8  & \underline{94.5}  & 91.3  & 90.8  & \textbf{94.6} \\
    \bottomrule
    \end{tabular}
  \label{visa_4shot_aupr}
\end{table*}

\begin{table*}[]
  \centering
  \caption{Subset-wise performance comparison results of the 1-shot setting for PRO on VisA.}
    \begin{tabular}{cccccccc}
    \toprule
    VisA  & \multicolumn{7}{c}{1-shot} \\
\cmidrule{2-8}    PRO   & SPADE & PatchCore & WinCLIP & APRIL-GAN & AnomalyGPT & PromptAD & KAG-prompt \\
    \midrule
    candle & 95.6  & 92.6  & 94.0    & 96.4  & 91.73 & 91.8  & 92.5 \\
    capsules & 83.1  & 66.6  & 73.6  & 88.4  & 82.41 & 70.0    & 78.9 \\
    cashew & 89.8  & 90.8  & 91.1  & 94.2  & 90.46 & 92.3  & 92.7 \\
    chewinggum & 73.9  & 78.2  & 91    & 92.1  & 87.31 & 89.8  & 88.2 \\
    fryum & 83.7  & 78.7  & 89.1  & 91.5  & 83.07 & 83.5  & 81.6 \\
    macaroni1 & 92.0    & 83.4  & 84.6  & 95.3  & 87.53 & 87.5  & 92.0 \\
    macaroni2 & 80.0    & 66.0    & 89.3  & 85.6  & 82.64 & 80.6  & 87.7 \\
    pcb1  & 81.3  & 79.0    & 82.5  & 90.4  & 83.08 & 89.2  & 85.8 \\
    pcb2  & 83.7  & 80.9  & 73.6  & 79.2  & 64.86 & 79.3  & 67.7 \\
    pcb3  & 84.3  & 78.1  & 79.5  & 81.3  & 69.91 & 84    & 75.7 \\
    pcb4  & 66.9  & 77.9  & 76.6  & 89.3  & 75.87 & 78.8  & 83.8 \\
    pipe fryum & 94.3  & 93.6  & 96.1  & 96.1  & 95.86 & 95.2  & 95.6 \\
    \midrule
    mean  & 84.1  & 80.5  & 85.1  & \textbf{90.0} & 82.9  & 85.1  & \underline{85.2} \\
    \bottomrule
    \end{tabular}%
  \label{visa_1shot_pro}
\end{table*}%

\begin{table*}[]
  \centering
  \caption{Subset-wise performance comparison results of the 2-shot setting for PRO on VisA.}
    \begin{tabular}{cccccccc}
    \toprule
    VisA  & \multicolumn{7}{c}{2-shot} \\
\cmidrule{2-8}    PRO   & SPADE & PatchCore & WinCLIP & APRIL-GAN & AnomalyGPT & PromptAD & KAG-prompt \\
    \midrule
    candle & 95.6  & 93.4  & 94.2  & 96.5  & 91.8  & 91.6  & 92.7 \\
    capsules & 85.4  & 67.9  & 75.9  & 89.1  & 85.7  & 70.8  & 85.9 \\
    cashew & 90.4  & 91.4  & 90.4  & 94.1  & 89.8  & 92.7  & 92.5 \\
    chewinggum & 73.8  & 78    & 90.9  & 92.1  & 85.4  & 87.8  & 88.0 \\
    fryum & 84.5  & 81.4  & 89.3  & 91.6  & 83.2  & 86.2  & 84.2 \\
    macaroni1 & 93.9  & 86.2  & 85.2  & 95.2  & 87.4  & 90.6  & 92.5 \\
    macaroni2 & 81.7  & 67.2  & 88.6  & 85.4  & 83.1  & 82.7  & 87.1 \\
    pcb1  & 87.2  & 86.1  & 83.8  & 90.6  & 82.9  & 90.2  & 87.0 \\
    pcb2  & 85.5  & 82.9  & 76.2  & 79.7  & 64.8  & 79.3  & 73.9 \\
    pcb3  & 86.1  & 82.2  & 82.3  & 81.5  & 72.4  & 84.7  & 75.5 \\
    pcb4  & 69.3  & 79.5  & 81.7  & 89.5  & 77.8  & 78.3  & 85.8 \\
    pipe fryum & 95.0    & 94.5  & 96.2  & 96.1  & 95.9  & 94.8  & 95.8 \\
    \midrule
    mean  & 85.7  & 82.6  & 86.2  & \textbf{90.1} & 83.4  & 85.8  & \underline{86.7} \\
    \bottomrule
    \end{tabular}%
  \label{visa_2shot_pro}%
\end{table*}%

\begin{table*}[]
  \centering
  \caption{Subset-wise performance comparison results of the 4-shot setting for PRO on VisA.}
    \begin{tabular}{cccccccc}
    \toprule
    VisA  & \multicolumn{7}{c}{4-shot} \\
\cmidrule{2-8}    PRO   & SPADE & PatchCore & WinCLIP & APRIL-GAN & AnomalyGPT & PromptAD & KAG-prompt \\
    \midrule
    candle & 95.7  & 94.1  & 94.4  & 96.3  & 93.1  & 90.6  & 93.4 \\
    capsules & 89.0    & 69.0    & 77.0    & 89.0    & 85.1  & 72.4  & 85.8 \\
    cashew & 90.4  & 92.1  & 91.3  & 94.1  & 90.8  & 92.8  & 92.1 \\
    chewinggum & 72.7  & 79.3  & 91.0    & 92.2  & 85.1  & 89.4  & 87.9 \\
    fryum & 86.2  & 81.0    & 89.7  & 91.5  & 83.1  & 80.3  & 84.6 \\
    macaroni1 & 95.1  & 89.6  & 86.8  & 95.2  & 90.7  & 91.5  & 93.7 \\
    macaroni2 & 86.0    & 78.3  & 90.5  & 85.8  & 84.9  & 87.2  & 87.6 \\
    pcb1  & 88.0    & 88.1  & 87.9  & 90.6  & 83.1  & 90.2  & 87.2 \\
    pcb2  & 87.0    & 83.7  & 78.0    & 80.2  & 71.3  & 76.3  & 77.7 \\
    pcb3  & 87.7  & 84.4  & 84.2  & 81.5  & 73.1  & 85.0    & 78.6 \\
    pcb4  & 74.7  & 83.5  & 84.2  & 89.7  & 79.1  & 83.4  & 87.0 \\
    pipe fryum & 95.0    & 95.0    & 96.6  & 95.9  & 95.4  & 95.3  & 95.9 \\
    \midrule
    mean  & 87.3  & 84.9  & 87.6  & \textbf{90.2} & 84.6  & 86.2  & \underline{87.6} \\
    \bottomrule
    \end{tabular}%
  \label{visa_4shot_pro}
\end{table*}

\clearpage

\begin{figure*}[]
    \centering
    \includegraphics[width=\linewidth]{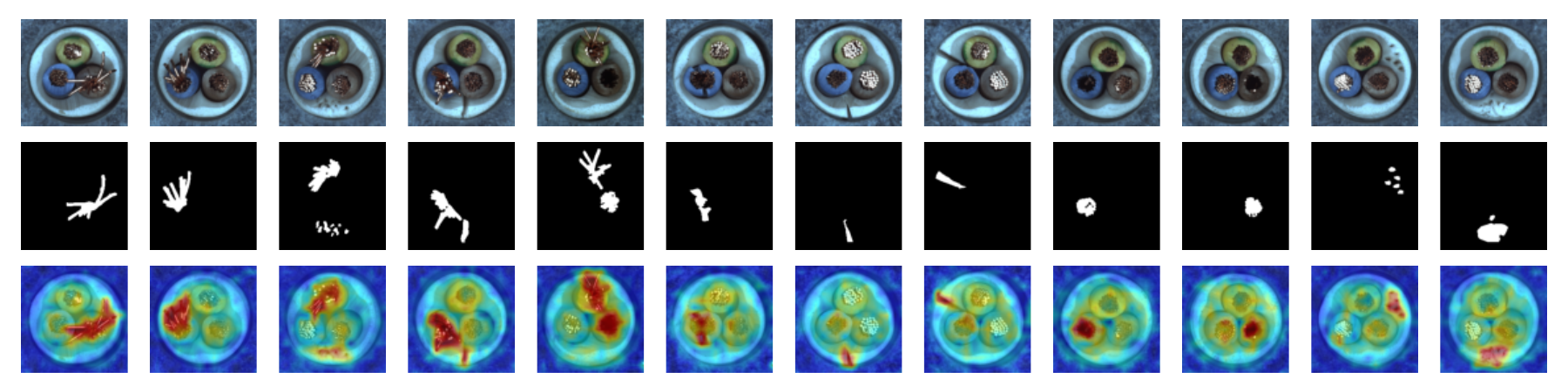}
    \caption{More visualization results for the data subset, cable, in MVTecAD.}
    \label{mvtec_cable}
\end{figure*}

\begin{figure*}[]
    \centering
    \includegraphics[width=\linewidth]{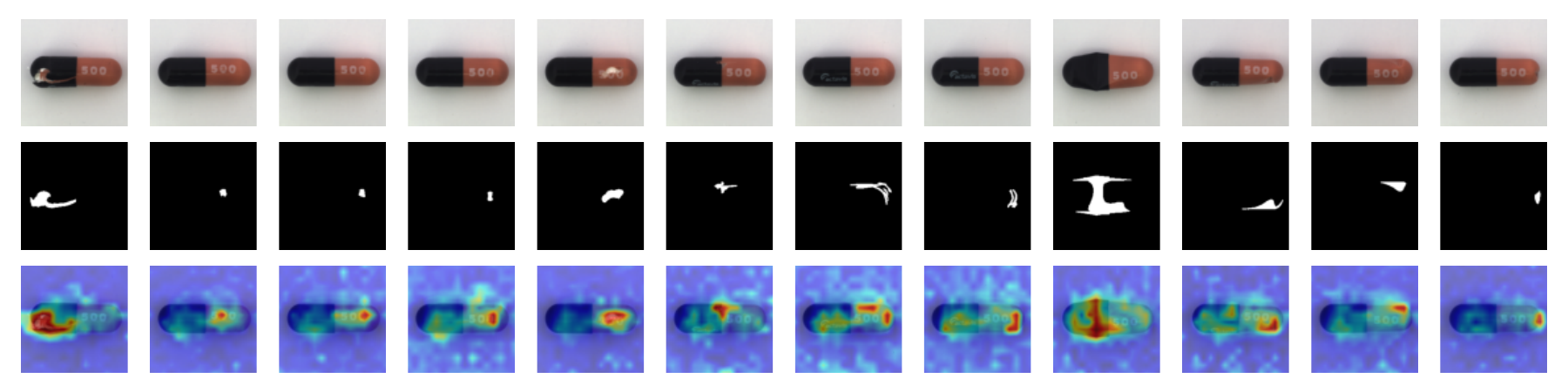}
    \caption{More visualization results for the data subset, capsule, in MVTecAD.}
    \label{mvtec_capsule}
\end{figure*}

\begin{figure*}[]
    \centering
    \includegraphics[width=\linewidth]{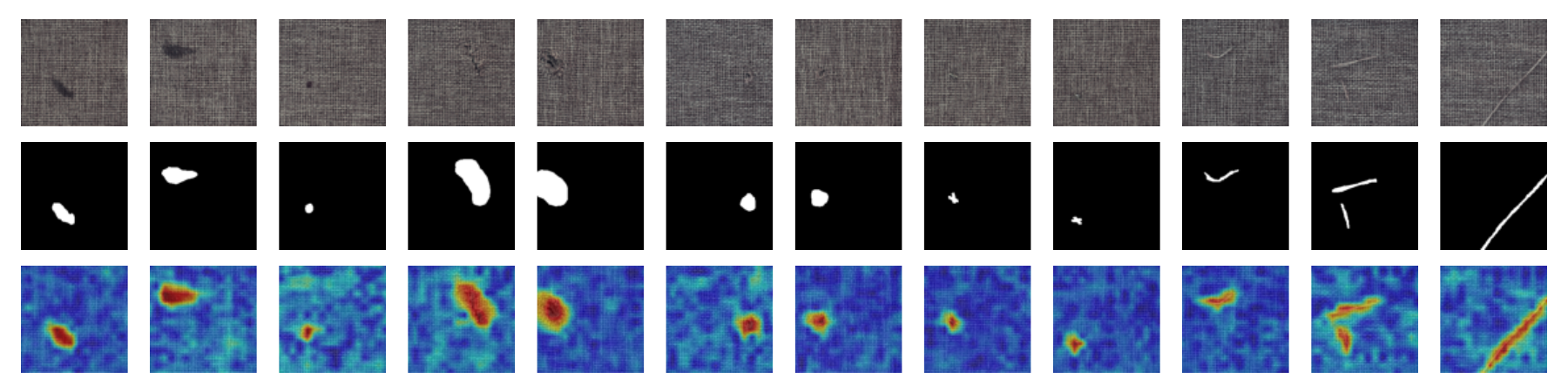}
    \caption{More visualization results for the data subset, carpet, in MVTecAD.}
    \label{mvtec_carpet}
\end{figure*}

\begin{figure*}[]
    \centering
    \includegraphics[width=\linewidth]{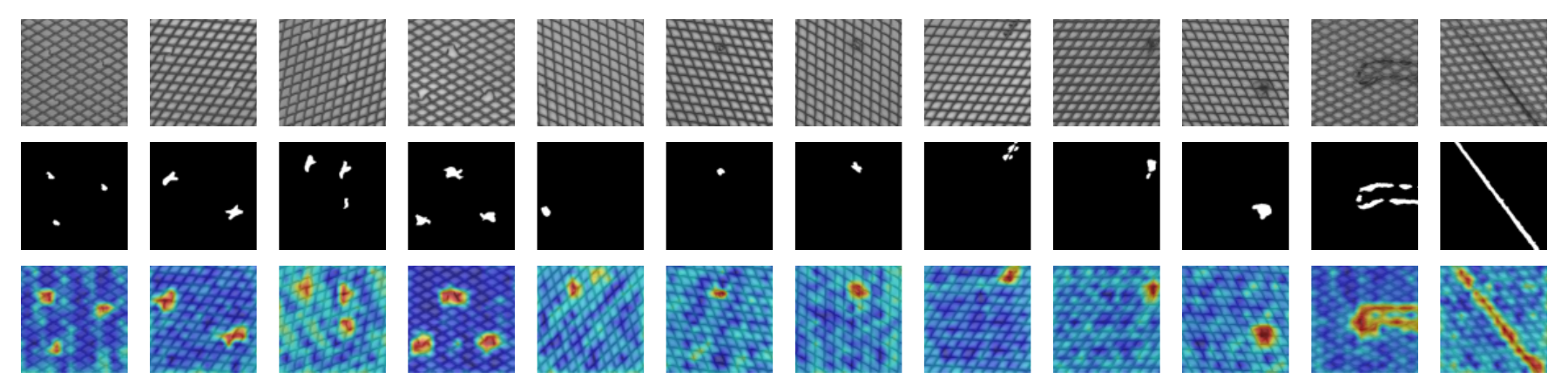}
    \caption{More visualization results for the data subset, grid, in MVTecAD.}
    \label{mvtec_grid}
\end{figure*}

\clearpage

\begin{figure*}[]
    \centering
    \includegraphics[width=\linewidth]{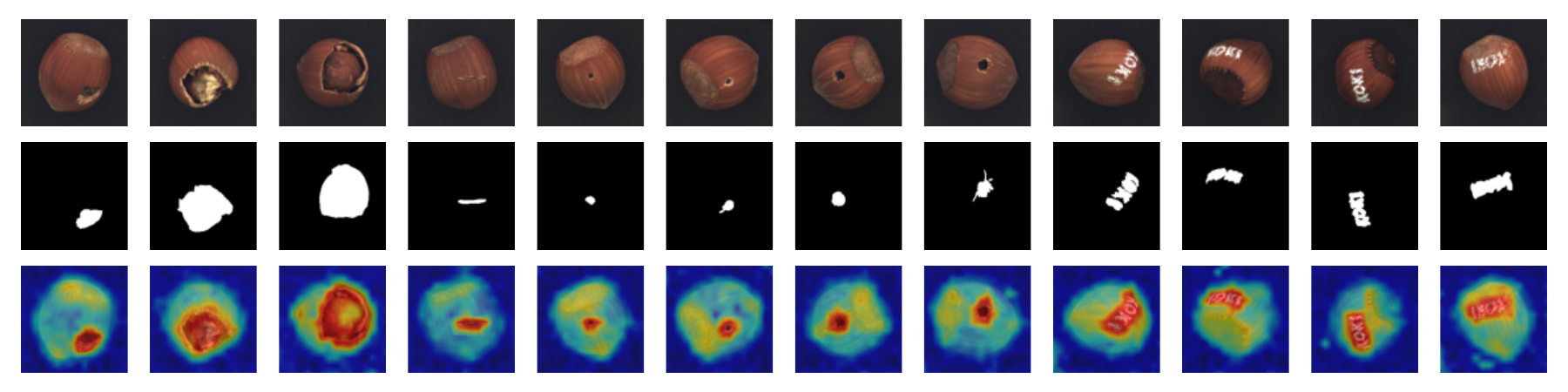}
    \caption{More visualization results for the data subset, hazelnut, in MVTecAD.}
    \label{mvtec_hazelnut}
\end{figure*}

\begin{figure*}[]
    \centering
    \includegraphics[width=\linewidth]{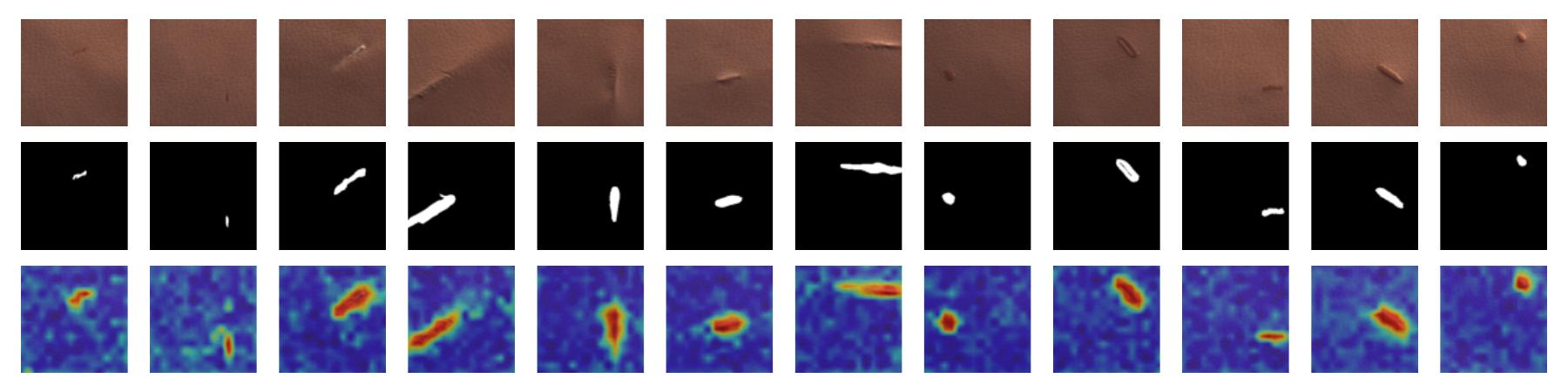}
    \caption{More visualization results for the data subset, leather, in MVTecAD.}
    \label{mvtec_leather}
\end{figure*}

\begin{figure*}[]
    \centering
    \includegraphics[width=\linewidth]{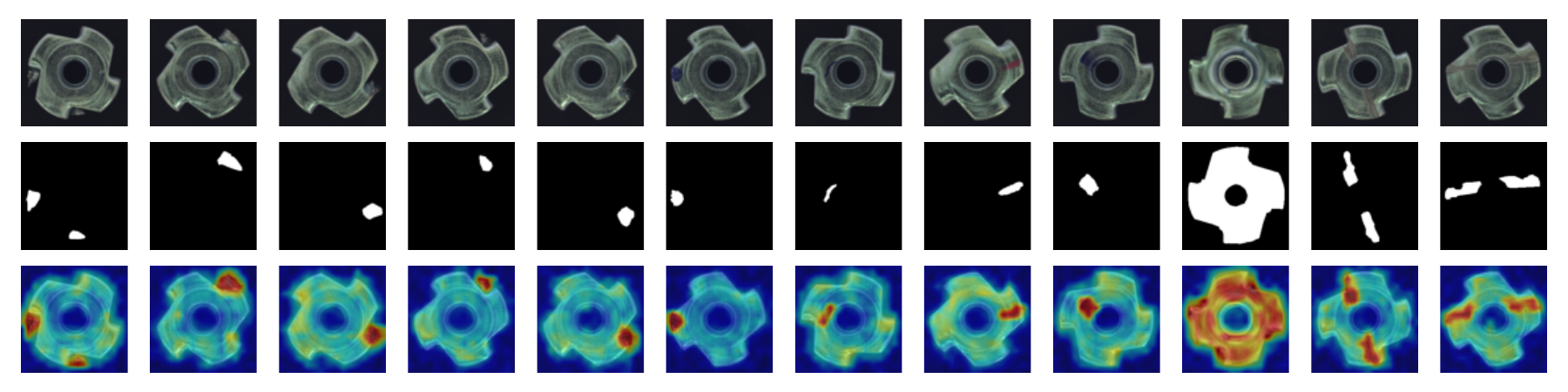}
    \caption{More visualization results for the data subset, metal nut, in MVTecAD.}
    \label{mvtec_metal nut}
\end{figure*}

\begin{figure*}[]
    \centering
    \includegraphics[width=\linewidth]{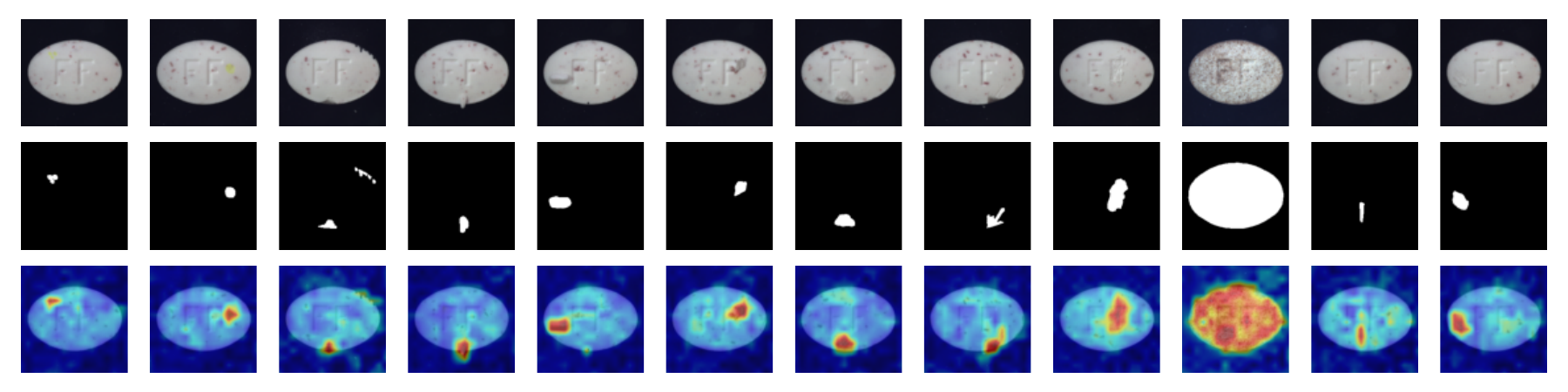}
    \caption{More visualization results for the data subset, pill, in MVTecAD.}
    \label{mvtec_pill}
\end{figure*}

\clearpage

\begin{figure*}[]
    \centering
    \includegraphics[width=\linewidth]{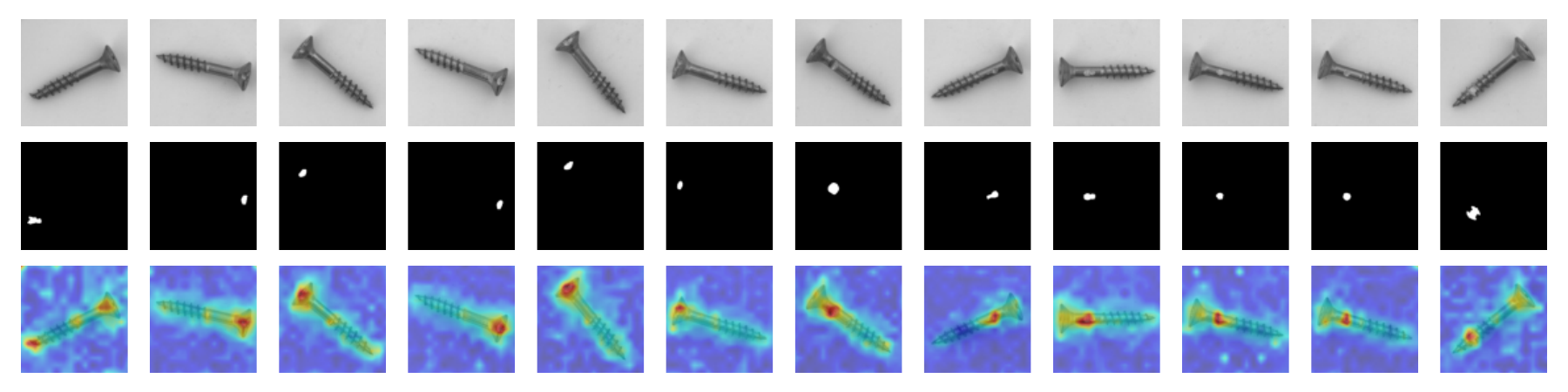}
    \caption{More visualization results for the data subset, screw, in MVTecAD.}
    \label{mvtec_screw}
\end{figure*}

\begin{figure*}[]
    \centering
    \includegraphics[width=\linewidth]{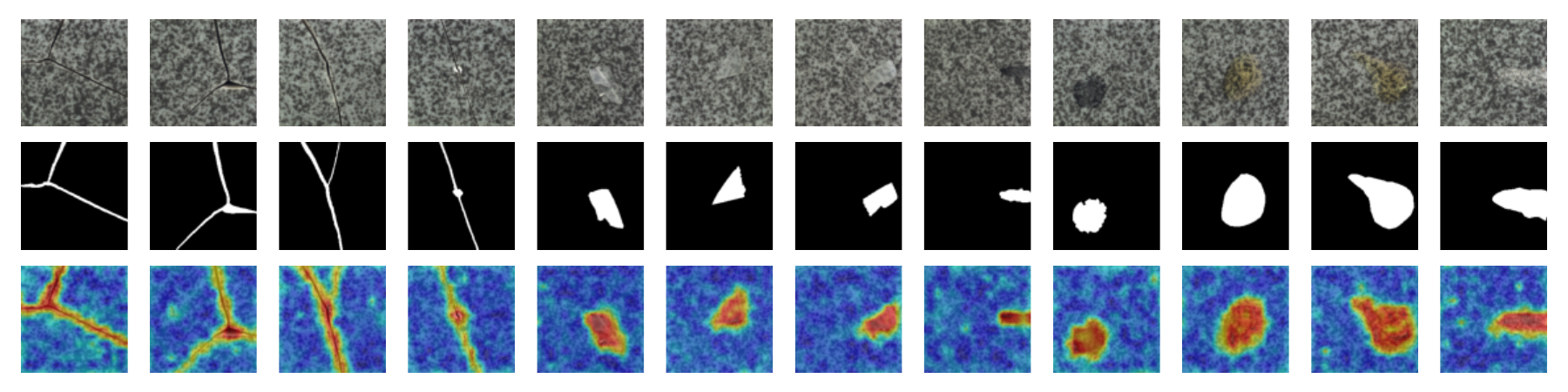}
    \caption{More visualization results for the data subset, tile, in MVTecAD.}
    \label{mvtec_tile}
\end{figure*}

\begin{figure*}[]
    \centering
    \includegraphics[width=\linewidth]{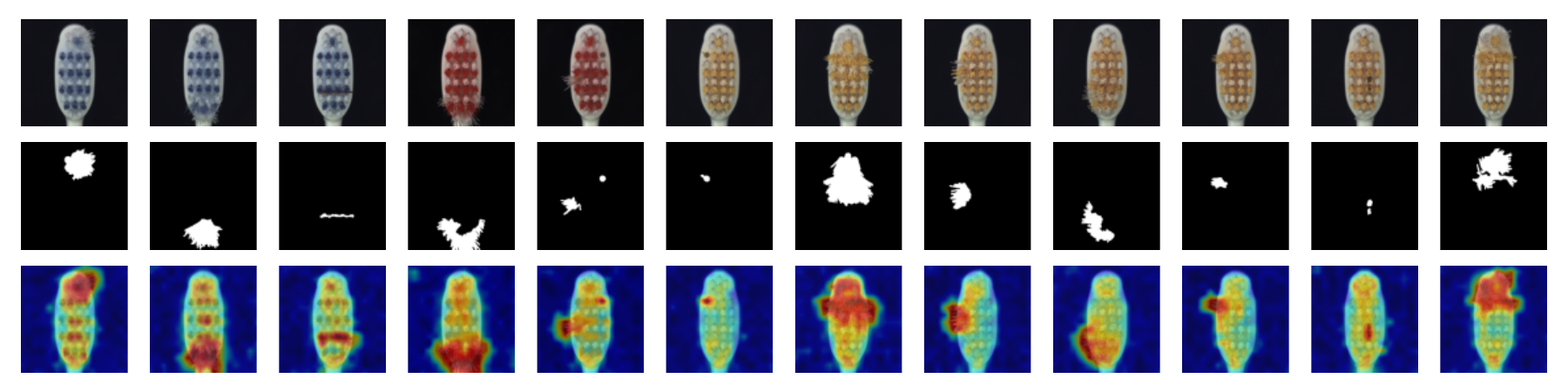}
    \caption{More visualization results for the data subset, toothbrush, in MVTecAD.}
    \label{mvtec_toothbrush}
\end{figure*}

\begin{figure*}[]
    \centering
    \includegraphics[width=\linewidth]{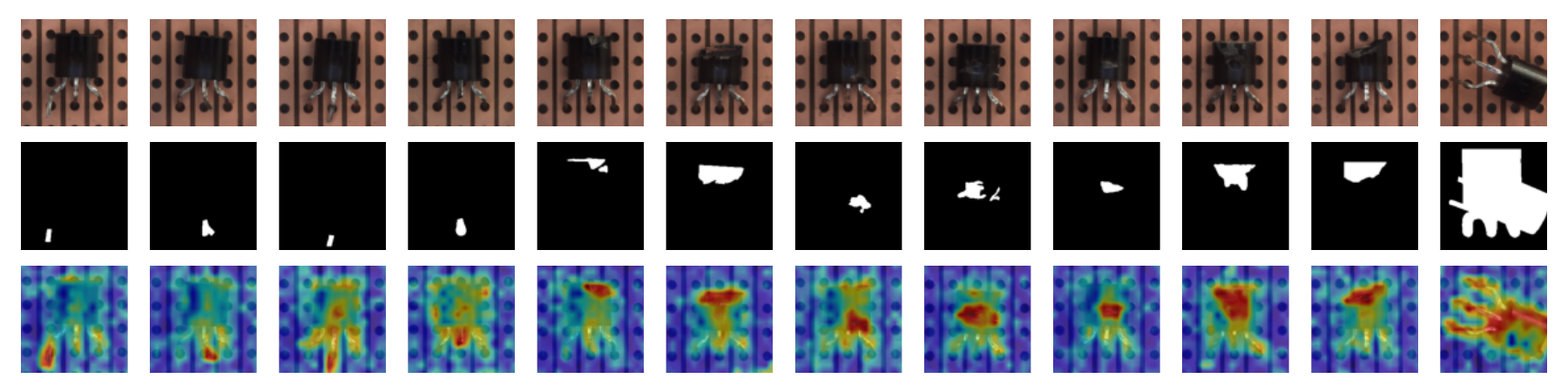}
    \caption{More visualization results for the data subset, transistor, in MVTecAD.}
    \label{mvtec_transistor}
\end{figure*}

\clearpage

\begin{figure*}[]
    \centering
    \includegraphics[width=\linewidth]{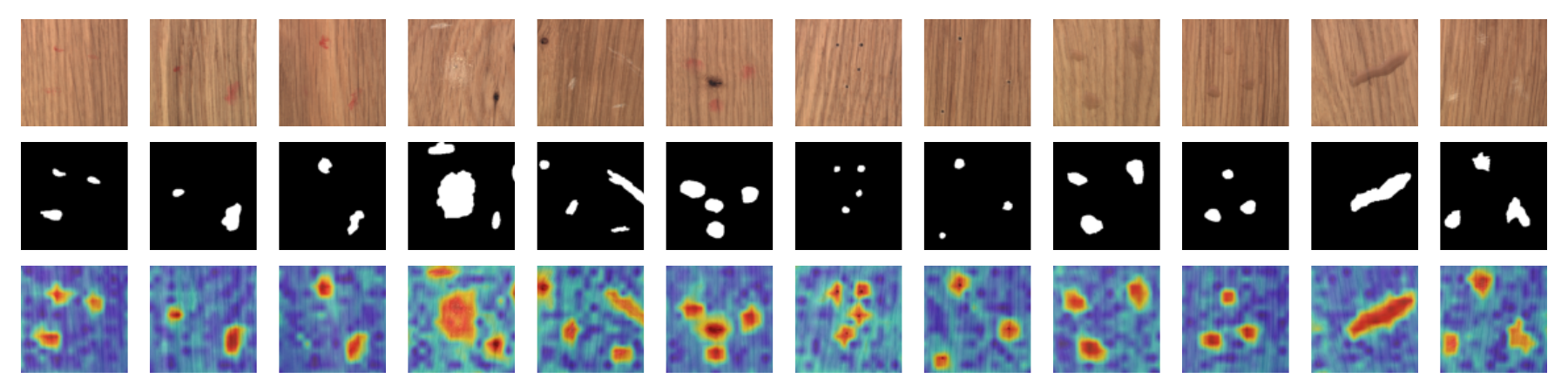}
    \caption{More visualization results for the data subset, wood, in MVTecAD.}
    \label{mvtec_wood}
\end{figure*}

\begin{figure*}[]
    \centering
    \includegraphics[width=\linewidth]{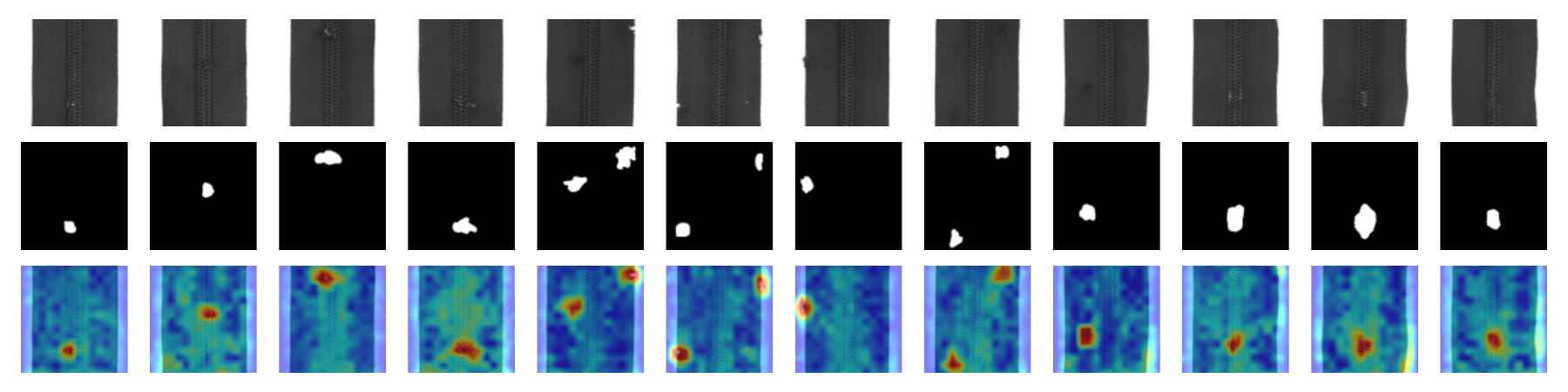}
    \caption{More visualization results for the data subset, zipper, in MVTecAD.}
    \label{mvtec_zipper}
\end{figure*}

\begin{figure*}[]
    \centering
    \includegraphics[width=\linewidth]{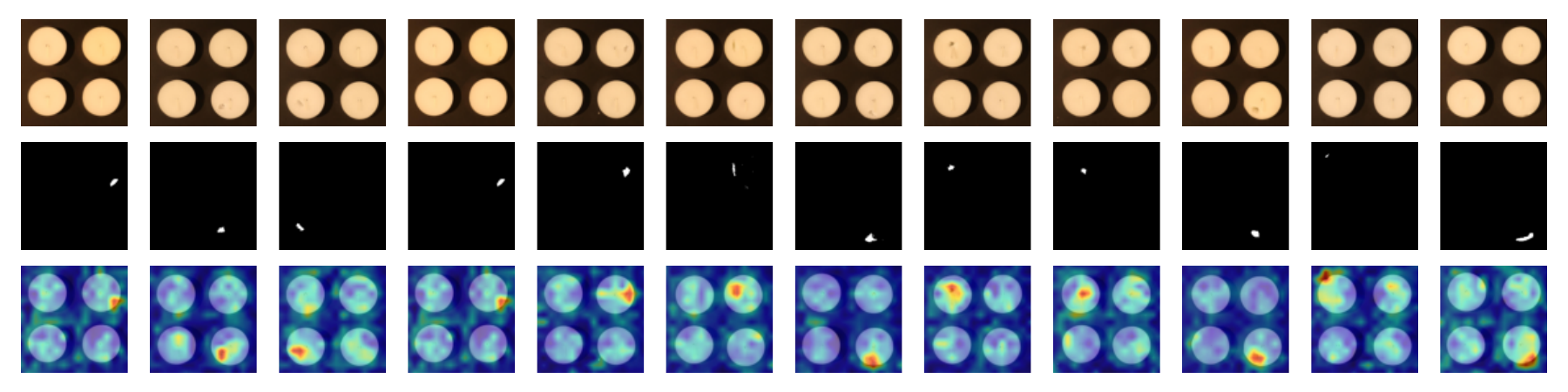}
    \caption{More visualization results for the data subset, candle, in VisA.}
    \label{visa_candle}
\end{figure*}

\begin{figure*}[]
    \centering
    \includegraphics[width=\linewidth]{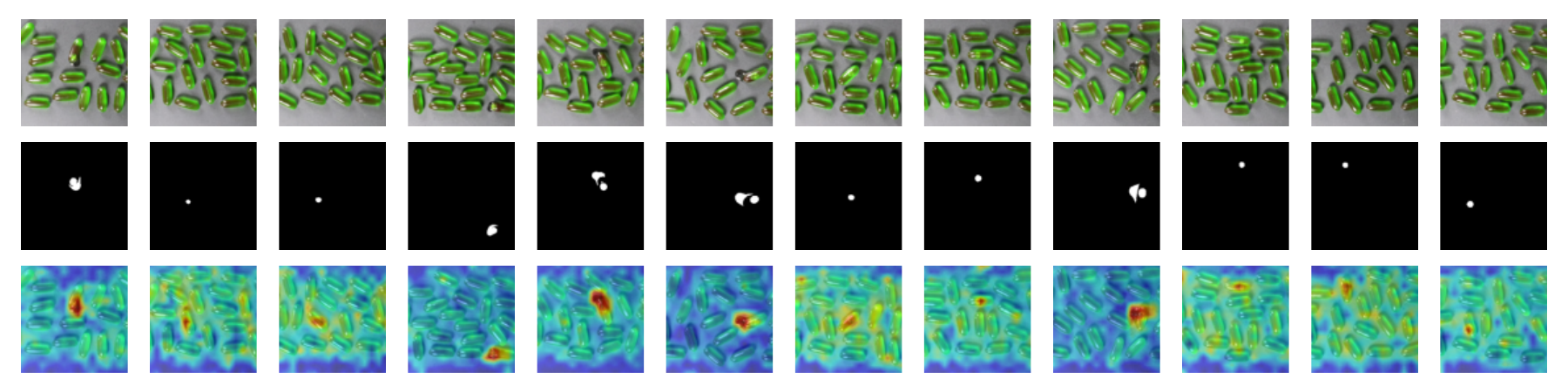}
    \caption{More visualization results for the data subset, capsules, in VisA.}
    \label{visa_capsules}
\end{figure*}

\clearpage

\begin{figure*}[]
    \centering
    \includegraphics[width=\linewidth]{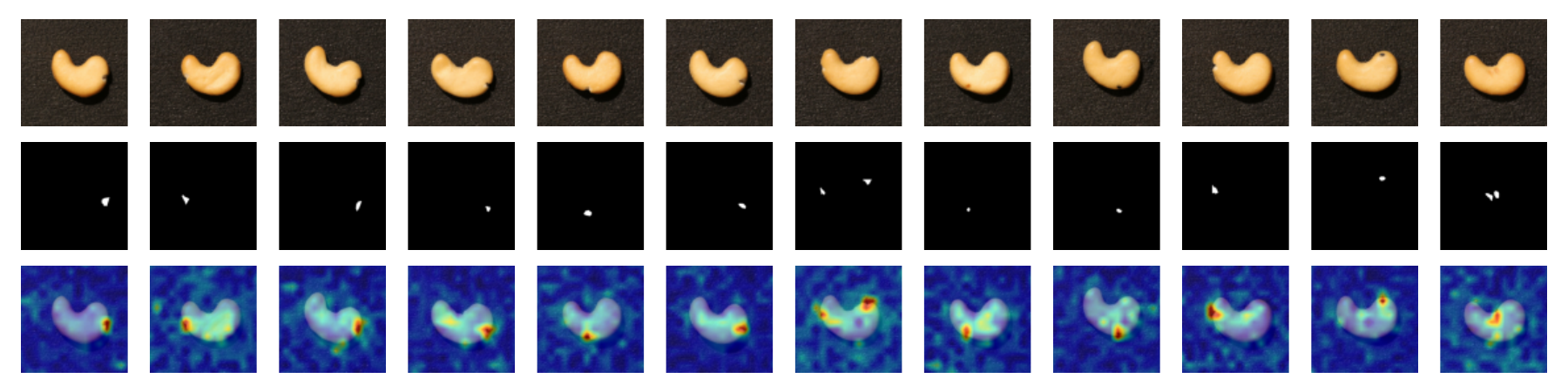}
    \caption{More visualization results for the data subset, cashew, in VisA.}
    \label{visa_cashew}
\end{figure*}

\begin{figure*}[]
    \centering
    \includegraphics[width=\linewidth]{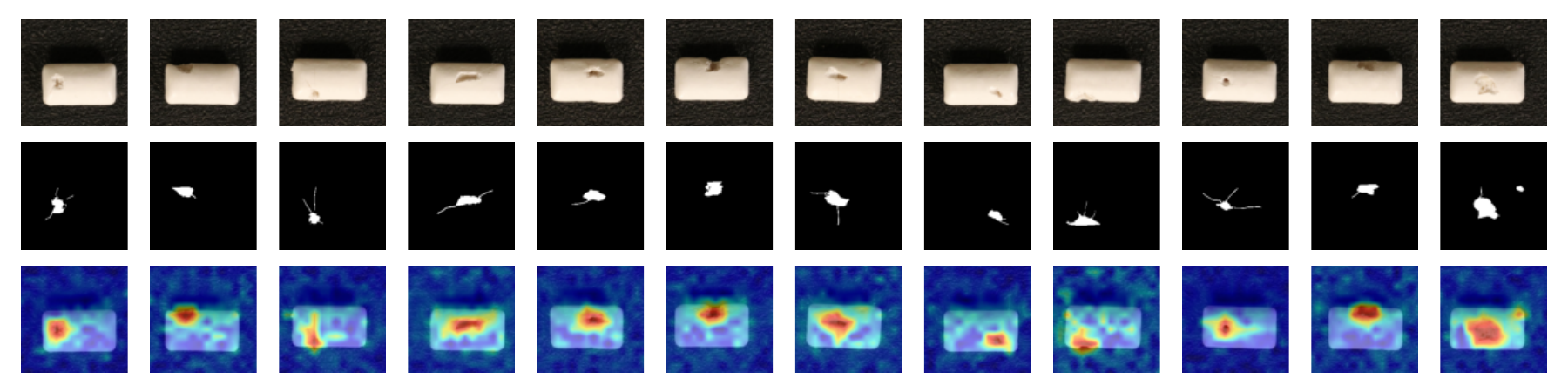}
    \caption{More visualization results for the data subset, chewinggum, in VisA.}
    \label{visa_chewinggum}
\end{figure*}

\begin{figure*}[]
    \centering
    \includegraphics[width=\linewidth]{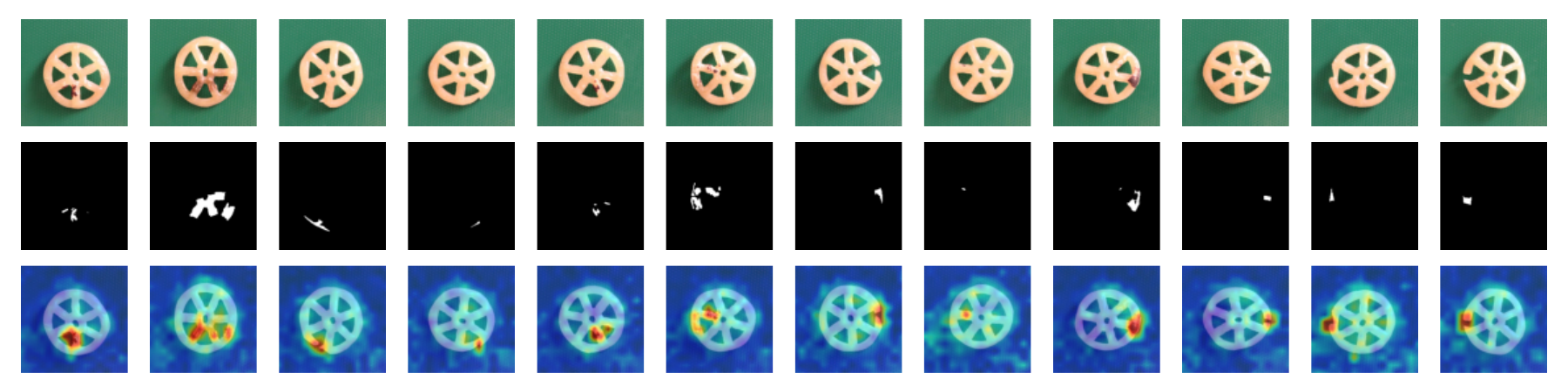}
    \caption{More visualization results for the data subset, fryum, in VisA.}
    \label{visa_fryum}
\end{figure*}

\begin{figure*}[]
    \centering
    \includegraphics[width=\linewidth]{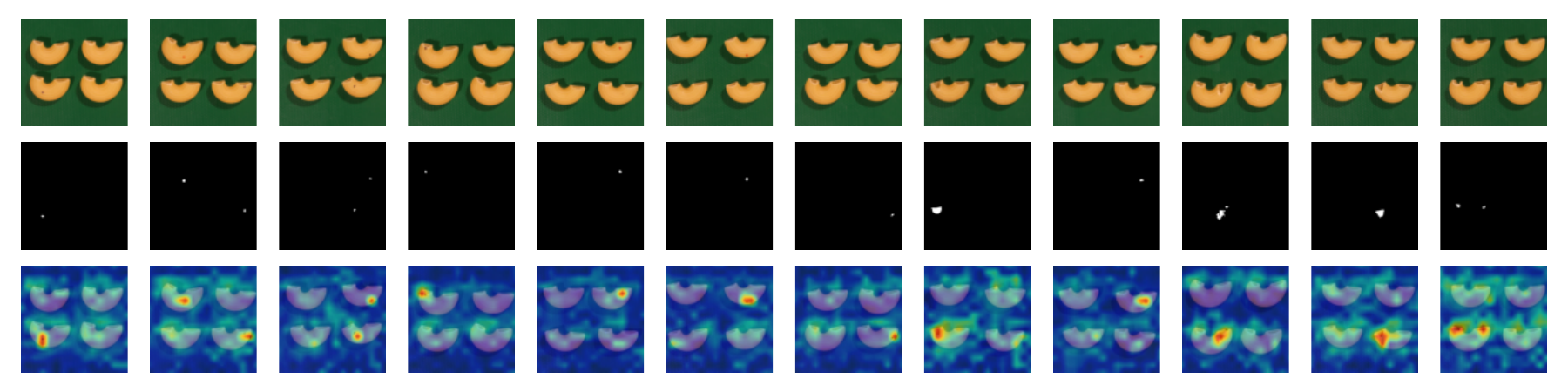}
    \caption{More visualization results for the data subset, macaroni1, in VisA.}
    \label{visa_macaroni1}
\end{figure*}

\clearpage

\begin{figure*}[]
    \centering
    \includegraphics[width=\linewidth]{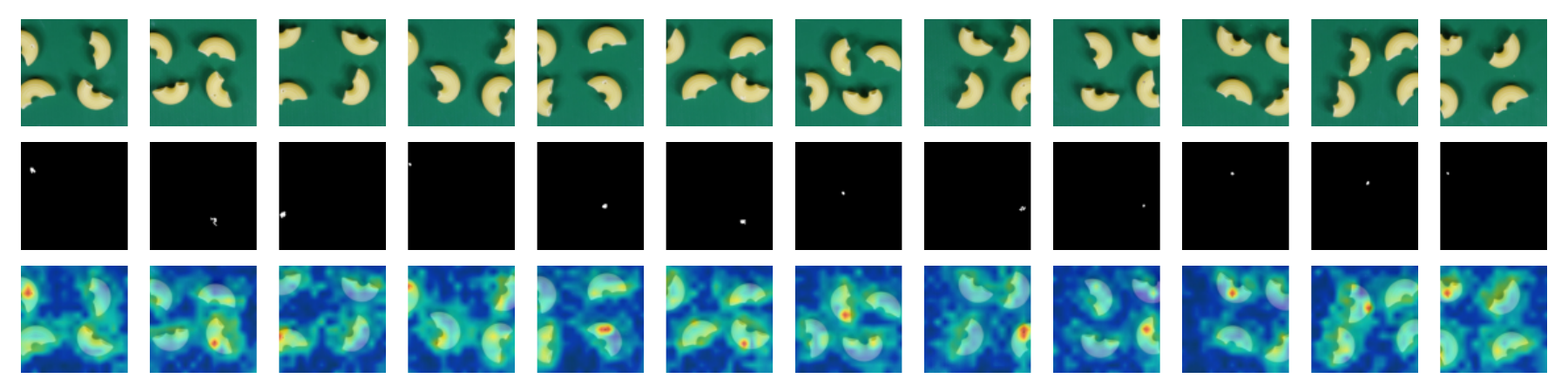}
    \caption{More visualization results for the data subset, macaroni2, in VisA.}
    \label{visa_macaroni2}
\end{figure*}

\begin{figure*}[]
    \centering
    \includegraphics[width=\linewidth]{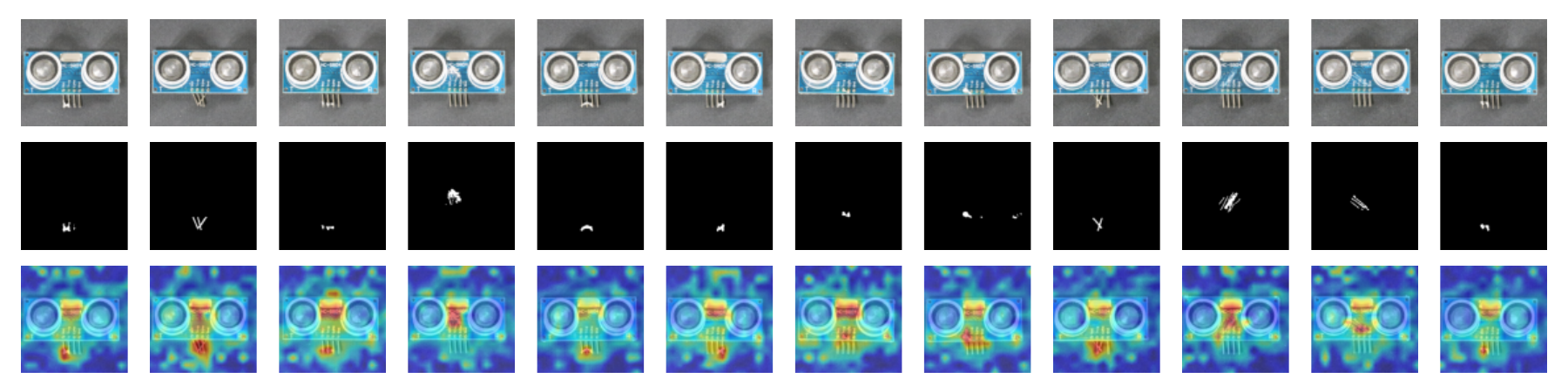}
    \caption{More visualization results for the data subset, pcb1, in VisA.}
    \label{visa_pcb1}
\end{figure*}

\begin{figure*}[]
    \centering
    \includegraphics[width=\linewidth]{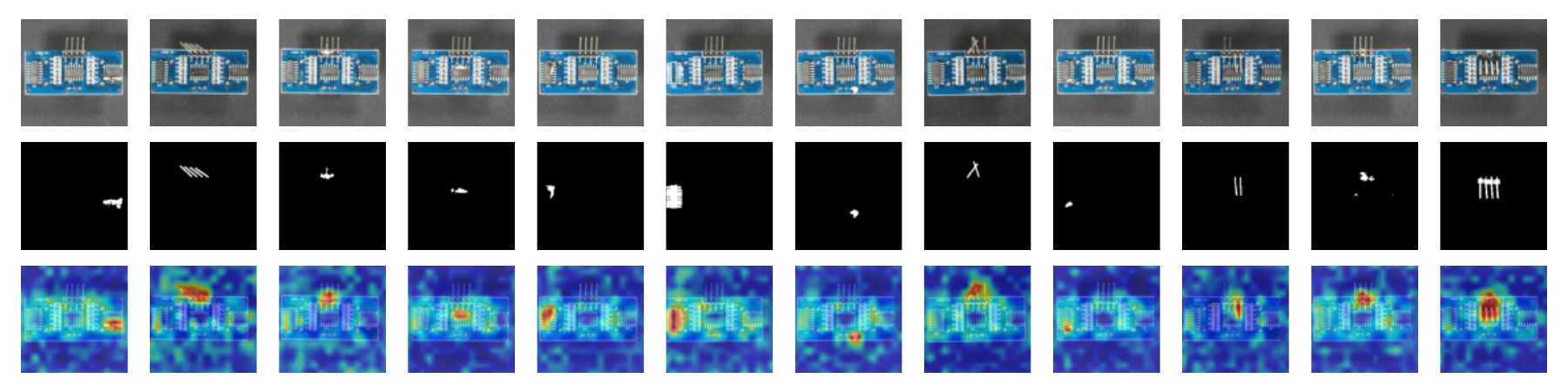}
    \caption{More visualization results for the data subset, pcb2, in VisA.}
    \label{visa_pcb2}
\end{figure*}

\begin{figure*}[]
    \centering
    \includegraphics[width=\linewidth]{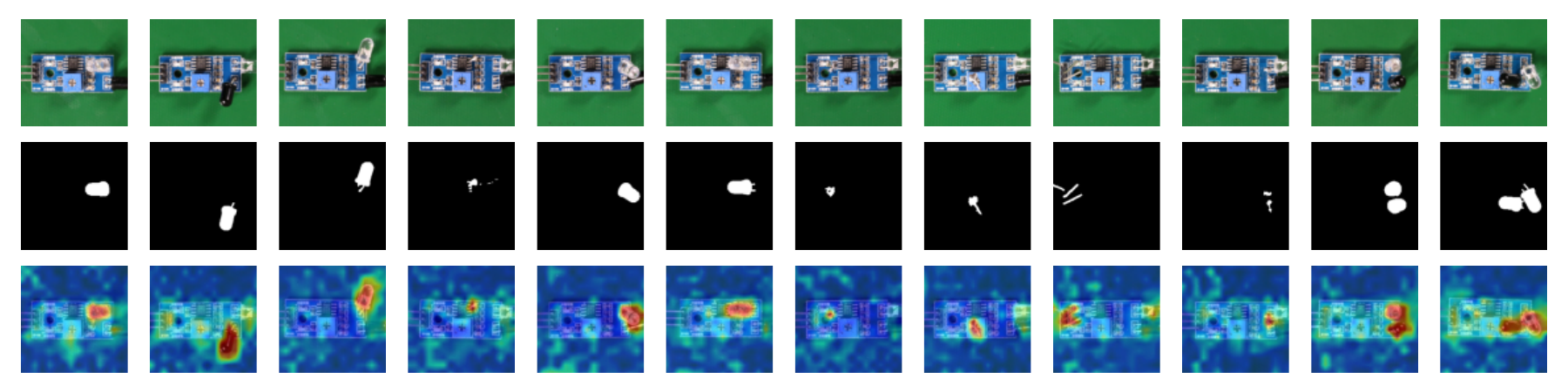}
    \caption{More visualization results for the data subset, pcb3, in VisA.}
    \label{visa_pcb3}
\end{figure*}

\clearpage

\begin{figure*}[]
    \centering
    \includegraphics[width=\linewidth]{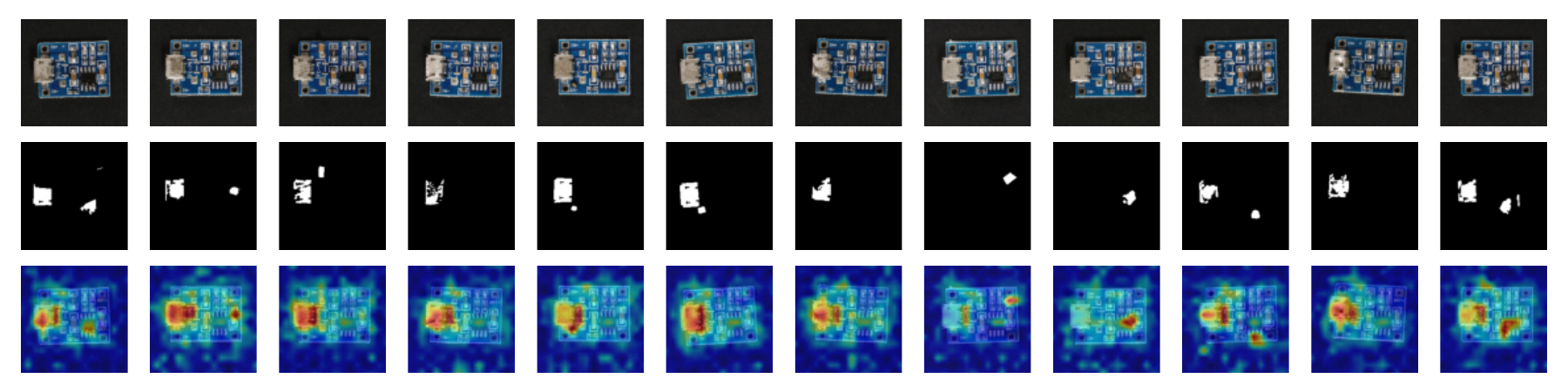}
    \caption{More visualization results for the data subset, pcb4, in VisA.}
    \label{visa_pcb4}
\end{figure*}

\begin{figure*}[]
    \centering
    \includegraphics[width=\linewidth]{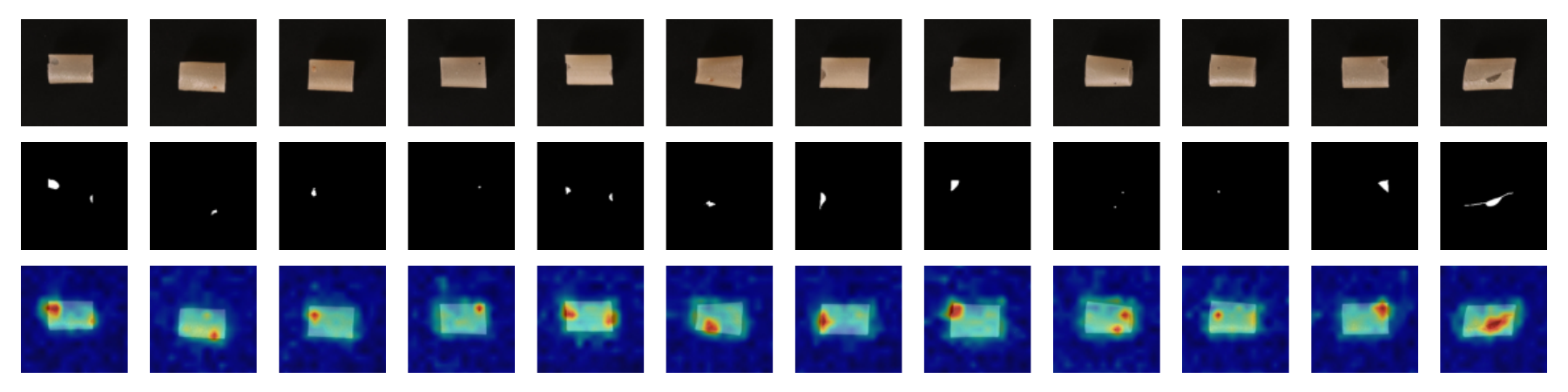}
    \caption{More visualization results for the data subset, pipe fryum, in VisA.}
    \label{visa_pipe_fryum}
\end{figure*}

\clearpage

\end{document}